\pdfoutput=1  %

\documentclass[10pt,twocolumn,letterpaper]{article}

\usepackage[pagenumbers]{cvpr}

\usepackage{xspace}
\usepackage{booktabs}
\usepackage{amsmath,amssymb,amsthm}
\usepackage{dsfont}            %
\usepackage{multirow}
\DeclareMathOperator*{\argmin}{arg\,min}

\usepackage{algorithm}
\usepackage{algpseudocode}
\usepackage{tikz}
\usetikzlibrary{arrows.meta,calc}
\definecolor{hgorange}{RGB}{222,110,30}  %
\definecolor{fmblue}{rgb}{0.21,0.49,0.74}
\definecolor{teamgreen}{RGB}{60,140,80}
\usepackage{colortbl}
\usepackage{adjustbox}
\colorlet{oursbg}{fmblue!9}                          %
\newcommand{\oursrow}{\rowcolor{oursbg}}             %
\newcommand{\venue}[1]{\textcolor{black!55}{#1}}     %
\newif\ifappendixmode
\DeclareRobustCommand{\mainpaper}[2]{\ifappendixmode\cref{#1}\else#2 of the main paper\fi}
\DeclareRobustCommand{\suppapp}[2]{\ifappendixmode#2\else#1\fi}
\newcommand{\appscale}[1]{\ifappendixmode#1\else1\fi}
\newcommand{\arxivcite}[2]{\ifappendixmode\cite{#1,#2}\else\cite{#1}\fi}

\definecolor{cvprblue}{rgb}{0.21,0.49,0.74}
\usepackage[pagebackref,breaklinks,colorlinks,allcolors=cvprblue]{hyperref}

\def\confName{CVPR}
\def\confYear{2027}

\newif\ifwithsupp
\withsupptrue  %
\ifwithsupp\appendixmodetrue\fi  %

\title{Groups Shape Futures, Futures Define Groups: Hypothesis-Conditioned Hypergraph Flow Matching for Collective Motion Forecasting}

\author{Rui Gan$^{1}$, Jiaxi Liu$^{1}$, Zheng Li$^{1}$, Sixu Li$^{1}$, Zihao Li$^{2}$, Keshu Wu$^{1,*}$, Yang Zhou$^{3}$, Xiaopeng Li$^{1}$ \\
$^{1}$University of Wisconsin-Madison, $^{2}$Marquette University, $^{3}$Texas A\&M University}

\begin{document}
\maketitle
{\renewcommand{\thefootnote}{$*$}\footnotetext{Corresponding author.}}  %
{\renewcommand{\thefootnote}{}\footnotetext{Project page: \url{https://keshuw95.github.io/H2Flow/}}}  %
\begin{abstract}
Predicting collective motion requires anticipating both where agents will move and which agents will move together. Existing group-aware forecasters infer their interaction structure from the observed motion before generating future hypotheses, even though different futures may involve different groups. We introduce hypothesis-conditioned groups, in which each predicted future forms its own groups as it is generated. Our framework, H\textsuperscript{2}Flow, integrates this principle into multi-hypothesis flow matching. A scene hypergraph encoder captures higher-order interactions in observed motion, while each hypothesis constructs its own multi-scale hypergraphs from its evolving estimate of the future. These groups guide trajectory generation through group tokens and a group-factorized prediction head, coupling the evolution of motion and interaction structure. To promote collective consistency, a hyperedge-level winner-take-all objective selects a shared winning hypothesis for interacting agents without group annotations. Experiments on basketball, American football, and soccer datasets demonstrate improved trajectory accuracy and collective forecasting performance. Analyses confirm that each hypothesis forms the groups of its own predicted future and that the gains come from groups formed from the predicted futures and from scoring the hypotheses on them.
\end{abstract}

\section{Introduction}
\label{sec:intro}
\begin{figure}[t]
  \centering
  \includegraphics[width=\linewidth]{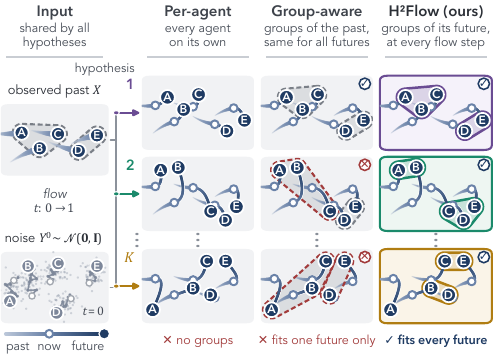}
  \caption{Hypothesis-conditioned groups: every hypothesis of H\textsuperscript{2}Flow forms the groups of the future it predicts.}
  \label{fig:teaser}
\end{figure}

Collective motion, from players running a set play to pedestrians crossing a square and vehicles merging into traffic, emerges from agents that coordinate in groups through higher-order interactions~\cite{helbing2000panic,couzin2005leadership,nagy2010hierarchical,battiston2021physics,zhang2023higher}. Forecasting it is central to sports analytics~\cite{xu2022groupnet,mao2023led}, social robot navigation, and autonomous driving~\arxivcite{shi2022mtr,ngiam2022scenetransformer}{gan2025goal,wu2025ai2,li2024sequencing, gan2025letspi}, where decisions depend on how groups act together. Because the future is uncertain, forecasters predict several plausible futures, each of which must capture both where every agent goes and which agents move together. Even for the same observed motion, group membership can differ across plausible futures and change over time within each future.

Recent generative forecasters represent the future as a set of $K$ hypotheses drawn from a learned distribution. Diffusion models~\cite{gu2022mid,mao2023led} and, more recently, flow matching models such as MoFlow~\cite{fu2025moflow} generate hypotheses for all agents jointly. Group-aware forecasters model interactions with hypergraphs whose hyperedges connect several agents at once~\cite{xu2022groupnet,xu2022dyngroupnet,li2022evolvehypergraph,lee2024mart}. Several such methods infer their hyperedges before the futures are generated, from the observed motion~\cite{xu2022groupnet,lee2024mart} or from a first-stage prediction~\cite{wu2024rhino}. All hypotheses therefore share the same groups, even when the predicted trajectories bring different agents together. Other methods update the interaction structure as each future unfolds, but model these interactions pairwise~\cite{li2020evolvegraph,liu2024betop}.

Even when a model predicts all agents jointly, a per-agent winner-take-all loss~\cite{guzmanrivera2012mcl} lets each agent select its own best hypothesis. In basketball, one hypothesis may predict the ball handler accurately and another the defender, even if neither captures their movements together. Low per-agent errors can therefore hide incoherent joint futures~\cite{weng2023joint}. Selecting one winner per scene during training can also be too coarse when a scene contains several loosely coupled groups. A group-level objective lets each group select a hypothesis that predicts its members well together.

We propose \textbf{hypothesis-conditioned groups}, in which each hypothesis forms its own groups as its future trajectories are generated. Our model, \textbf{H\textsuperscript{2}Flow}, couples trajectory prediction and group formation in two directions (\cref{fig:teaser}). \emph{Futures define groups.} Flow matching with data prediction gives each hypothesis an estimate of its future at every flow step. Self-conditioning~\cite{chen2023analogbits} passes that estimate to the next step, where each decoder block rebuilds hyperedges of several sizes using the agent tokens and the distances between agents in the estimated future. This reuses the previous estimate without an additional network evaluation. \emph{Groups shape futures.} Agents attend to their hypothesis's group tokens, and a group-factorized endpoint adds a trajectory correction shared by each group's members. During training, a hyperedge-level winner-take-all loss compares hypotheses by the mean trajectory error over the members each assigns to a given hyperedge. On NBA, the groups formed by each hypothesis more closely match those in its own predicted future than those in another hypothesis's future (\cref{sec:analysis}). Removing the hyperedge-level loss increases scene-level trajectory error (\cref{sec:ablation}).

We make three contributions.
\begin{enumerate}
  \item \textbf{Futures Define Groups.} We introduce hypothesis-conditioned groups that update as each hypothesis refines its trajectory estimates.
  \item \textbf{Groups Shape Futures.} Group tokens and a group-factorized endpoint couple the predictions of interacting agents. A hyperedge-level winner-take-all loss selects a shared winner per group without group labels.
  \item \textbf{Accurate and Coherent Futures.} At a 4.0\,s horizon, H\textsuperscript{2}Flow achieves the lowest trajectory errors among the methods evaluated on NBA and NFL at every level from single agents to whole scenes. On NBA, it reduces per-agent min$_{20}$ADE by 3.7\% and Scene-min$_{20}$ADE by 6.2\% relative to MoFlow.
\end{enumerate}

\section{Related Work}
\label{sec:related}

\noindent\textbf{Multi-agent Trajectory Forecasting.}
Early forecasters model social interactions by pooling over neighbors~\cite{gupta2018socialgan} or with spatio-temporal graphs~\arxivcite{salzmann2020trajectron}{xu2023eqmotion}, and transformers attend densely across agents and time~\cite{ngiam2022scenetransformer,shi2022mtr}.
Generative models capture the multi-modality of future motion with diffusion~\cite{gu2022mid,mao2023led,jiang2023motiondiffuser,capellera2025u2diff, you2025followgen} or flow matching~\cite{lipman2023flow,liu2023rectified,yan2025trajflow}: MoFlow~\cite{fu2025moflow} generates $K$ correlated scene-level hypotheses and distills them into a one-step student~\cite{li2018imle}.
Self-conditioning~\cite{chen2023analogbits,yan2025trajflow} feeds a generative model its own previous clean estimate; in H\textsuperscript{2}Flow, it is the future from which each hypothesis forms its groups.

\noindent\textbf{Group-aware Relational Reasoning.}
Hypergraph neural networks~\cite{feng2019hgnn} pass messages over hyperedges that connect any number of nodes~\cite{zhou2006hypergraph}, a natural fit for the higher-order interactions of groups~\cite{battiston2021physics,zhang2023higher}.
Group-aware forecasters pool agents into learned groups~\cite{bae2022gpgraph} or build such hypergraphs over them: GroupNet~\cite{xu2022groupnet} at multiple scales from pairwise affinities, MART~\cite{lee2024mart} with an adaptive group estimator and a hyper-relational transformer, DynGroupNet~\cite{xu2022dyngroupnet} and EvolveHypergraph~\cite{li2022evolvehypergraph} with groups that evolve over the horizon, and RHINO~\cite{wu2024rhino} over the predicted modes of a first-stage forecaster.
GroupNet, MART, and RHINO infer their hyperedges once, before the futures are sampled, so every sampled future shares the same groups.
Structure that does follow individual futures has, to our knowledge, been pairwise: EvolveGraph~\cite{li2020evolvegraph} re-encodes an interaction graph along each sampled rollout, BeTopNet~\cite{liu2024betop} selects the agents that each mode query attends to in every decoder layer, and M2I~\cite{sun2022m2i} and GameFormer~\cite{huang2023gameformer} condition the prediction of an agent on the predicted futures of others.
Hypothesis-conditioned groups bring this per-future structure to hyperedges: each hypothesis of H\textsuperscript{2}Flow forms its own hyperedges from its estimated future at every flow step.

\noindent\textbf{Multi-hypothesis Learning and Joint Evaluation.}
Winner-take-all objectives~\cite{xu2025awta} build on multiple choice learning~\cite{guzmanrivera2012mcl}; in multi-agent scenes, the winner is selected per agent (marginal) or per scene (joint)~\cite{ngiam2022scenetransformer}, and marginal and joint metrics can rank methods differently~\cite{weng2023joint}.
H\textsuperscript{2}Flow adds an intermediate level that selects one winner per hyperedge, scoring each hypothesis on the groups it formed, and we evaluate it per agent, group, and scene.

\section{Method}
\label{sec:method}

\subsection{Problem Formulation}
\label{sec:background}
\label{sec:bg_motion}
A scene has $N$ agents observed over $T_p$ frames.
For every agent and frame, the context $C\in\mathbb{R}^{N\times T_p\times 6}$ holds the position, the position relative to the last observed frame, and the velocity.
The task is to forecast the future $Y\in\mathbb{R}^{N\times 2T_f}$: the positions of all agents over the next $T_f$ frames, relative to their last observed positions.
A multi-hypothesis forecaster outputs $K$ scene-level hypotheses $S_k\in\mathbb{R}^{N\times 2T_f}$, each a joint future of all agents with row $S_{k,i}$ for agent $i$, and logits $\zeta_k\in\mathbb{R}^{N}$ that score hypothesis $k$ for each agent.
Hypothesis $k$ errs on agent $i$ by its average displacement $d_{k,i}=\frac{1}{T_f}\sum_{\tau=1}^{T_f}\lVert S_{k,i,\tau}-Y_{i,\tau}\rVert_2$.
A forecast is accurate per agent if every agent has some close hypothesis, and jointly if one hypothesis is close for all agents of a group (\cref{sec:data}).

\subsection{Preliminaries}
\label{sec:bg_moflow}
\noindent\textbf{Hypergraphs as Interaction Representations.}
\label{sec:bg_hypergraph}
A hypergraph $(\mathcal{V},\mathcal{E})$ over the agents $\mathcal{V}=\{1,\dots,N\}$ represents higher-order interactions~\cite{zhou2006hypergraph,feng2019hgnn,battiston2021physics}: each of its $E$ hyperedges is a group $e\subseteq\mathcal{V}$ of any size whose agents interact jointly, encoded by the incidence matrix $H\in\{0,1\}^{N\times E}$ with $H_{i,e}=\mathds{1}[i\in e]$ and $|e|=\sum_i H_{i,e}$.

\noindent\textbf{Flow Matching with Data Prediction.}
Flow matching~\cite{lipman2023flow,liu2023rectified} transports noise $Y^0\sim p_0=\mathcal{N}(\mathbf{0},\mathbf{I})$ to the future $Y^1=Y$ along a velocity field, $\mathrm{d}Y^t/\mathrm{d}t=v_\theta(Y^t,C,t)$ over the flow time $t\in[0,1]$, which it regresses on the linear path $Y^t=(1-t)Y^0+tY^1$:
\begin{equation}
  \mathcal{L}_{\mathrm{CFM}}=\mathbb{E}_{t,Y^0,Y^1}\big\lVert v_\theta(Y^t,C,t)-(Y^1-Y^0)\big\rVert_2^2.
  \label{eq:cfm}
\end{equation}
A data predictor $D_\theta$ estimates the clean future $Y^1$ and defines the velocity
\begin{equation}
  v_\theta(Y^t,C,t)=\frac{D_\theta(Y^t,C,t)-Y^t}{1-t},
  \label{eq:vel}
\end{equation}
so that \cref{eq:cfm} regresses $D_\theta$ toward $Y^1$ with weight $(1-t)^{-2}$.
Every step of an Euler solver thus yields an estimate of the clean future, and the last one is the forecast.

\noindent\textbf{Multi-hypothesis Flow Matching.}
For multi-modal futures, $D_\theta$ outputs $K$ hypotheses $S_k$ and logits $\zeta$ from one noisy input~\cite{fu2025moflow}, and each hypothesis follows its own velocity $v^{(k)}_\theta=(S_k-Y^t_k)/(1-t)$.
In training, the hypotheses share one noise draw, $Y^t_k=Y^t$, and differ through learned query embeddings; only the best hypothesis of each agent is regressed toward $Y^1$, without the weight $(1-t)^{-2}$,
\begin{equation}
  \mathcal{L}_{\mathrm{agent}}=\mathbb{E}_{t,Y^0,Y^1}\Big[\frac{1}{N}\sum_{i=1}^{N}\min_{k}\,d_{k,i}\Big],
  \label{eq:agent}
\end{equation}
and a cross-entropy $\mathcal{L}_{\mathrm{cls}}$ trains $\zeta$ to pick this winner; as in multiple choice learning~\cite{guzmanrivera2012mcl}, the hypotheses spread over the modes of the future.
This per-agent winner-take-all never asks one hypothesis to be right for several agents at once; the per-scene one, $\min_k\frac{1}{N}\sum_i d_{k,i}$, asks it for all agents, too coarse for loosely coupled groups.

\subsection{Model Structure}
\label{sec:overview}
\label{sec:hcg}
\begin{figure*}[!t]
  \centering
  \includegraphics[width=0.98\textwidth]{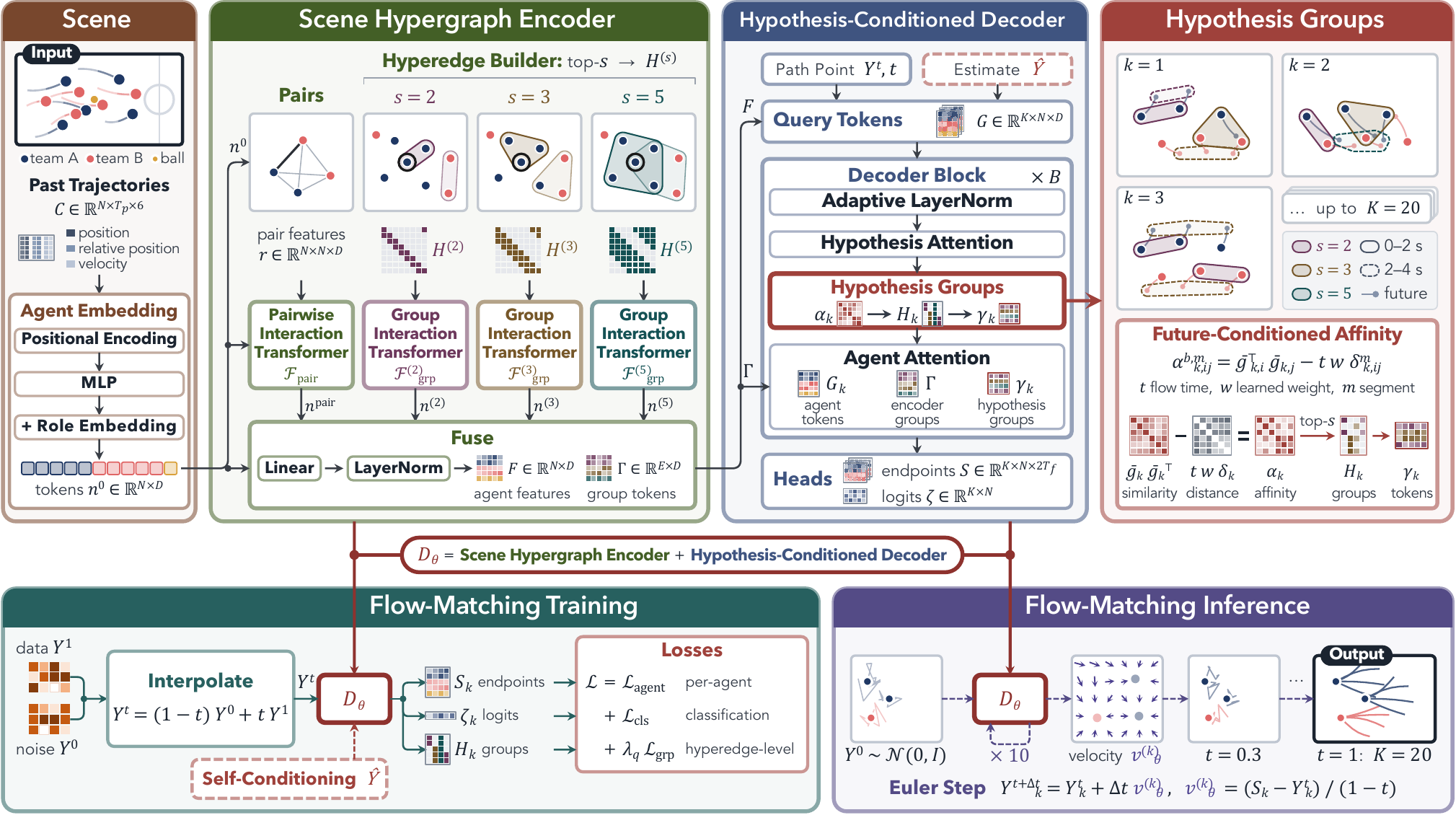}
  \caption{Overview of H\textsuperscript{2}Flow. The encoder runs once per scene; in every decoder block, each hypothesis forms the groups of its estimated future, and these groups shape its prediction.}
  \label{fig:overview}
\end{figure*}

Group-aware forecasters decode all $K$ hypotheses with one hypergraph per scene, built before the futures are generated~\cite{xu2022groupnet,lee2024mart,wu2024rhino}, yet which agents move together depends on which future unfolds, and it changes over the horizon.
We therefore propose hypothesis-conditioned groups: every hypothesis forms the groups of the future it predicts.
The data predictor $D_\theta$ of H\textsuperscript{2}Flow (\cref{fig:overview}) pairs a scene hypergraph encoder, run once per scene, with a hypothesis-conditioned decoder, run at every flow step:
\begin{align}
  (F,\Gamma) &= \mathrm{Enc}_\theta(C), \nonumber\\
  (S,\zeta,\mathcal{H}) &= \mathrm{Dec}_\theta(Y^t,t,\hat Y;\,F,\Gamma).
  \label{eq:factor}
\end{align}
The encoder gives agent features $F\in\mathbb{R}^{N\times D}$ and encoder group tokens $\Gamma\in\mathbb{R}^{E\times D}$; from the $K$ flow states $Y^t$ and their estimate $\hat Y\in\mathbb{R}^{K\times N\times 2T_f}$, the decoder predicts the endpoints $S$ (estimates of $Y^1$), logits $\zeta\in\mathbb{R}^{K\times N}$, and groups $\mathcal{H}=\{H_k\}_{k=1}^K$.
In the decoder, the estimated futures define the groups of each hypothesis, and these groups shape its future (\cref{sec:hypgroups}).
\suppapp{The supplementary material}{The appendix} details all layers.

\subsubsection{Scene Hypergraph Encoder}
\label{sec:encoder}
\noindent\textbf{Agent Embedding.}
The context frames of agent $i$, each with a sinusoidal encoding of its index $\tau$, are embedded and projected into $n^0_i$, with a role embedding (team or ball) added on NBA.

\noindent\textbf{Pairwise Interaction Transformer with Relative Geometry.}
Whether two agents interact depends on their relative position and motion, \eg, a defender closing in on a ball handler.
Every ordered pair therefore starts from a relation that adds this geometry to both embeddings,
\begin{equation}
  r_{ij}=\mathrm{FFN}_r\big([n^0_i;\,n^0_j]\big)+\mathrm{FFN}_g\big([x_j-x_i;\,\dot x_j-\dot x_i;\,\lVert x_j-x_i\rVert_2]\big),
  \label{eq:geo}
\end{equation}
where $x_i$ and $\dot x_i$ are the last observed position and velocity of agent $i$ and $[\cdot\,;\,\cdot]$ is concatenation.
A pairwise interaction transformer $\mathcal{F}_{\mathrm{pair}}$ of $L$ relational attention layers~\cite{lee2024mart} then lets each agent attend to all agents through its relations to them and updates the relations, giving $n^{\mathrm{pair}}_i=\mathcal{F}_{\mathrm{pair}}(n^0,r)_i$, where $n^0$ and $r$ stack all $n^0_i$ and $r_{ij}$.

\noindent\textbf{Hyperedge Builder.}
To model groups at several scales, we fix a set of group sizes $\mathcal{S}=\{s_1,\dots,s_{|\mathcal{S}|}\}$ and learn the members: the hyperedge of size $s\in\mathcal{S}$ anchored at agent $j$ holds $j$ and the $s-1$ agents of highest affinity to it, $H^{(s)}_{i,j}=\mathds{1}[\alpha_{ij}\ge\alpha_j^{(s)}]$, where $\alpha_{ij}=\bar z_i^{\top}\bar z_j$ is the cosine affinity of centered agent projections $\bar z_i$ and $\alpha_j^{(s)}$ is the $s$-th largest affinity of agent $j$.
Together, the $E=|\mathcal{S}|N$ hyperedges form $H=[H^{(s_1)},\dots,H^{(s_{|\mathcal{S}|})}]\in\{0,1\}^{N\times E}$.
A straight-through estimator~\cite{bengio2013estimating} with temperature $\kappa$ trains the members through the hard selection.
The decoder reuses this builder with a hypothesis-dependent affinity (\cref{sec:hypgroups}).

\noindent\textbf{Group Interaction Transformers.}
For every size $s$, a group interaction transformer $\mathcal{F}^{(s)}_{\mathrm{grp}}$ of $L$ layers~\cite{lee2024mart} lets each agent attend to all agents through its hyperedges and updates each hyperedge feature $h_e$ from its members, giving $n^{(s)}_i=\mathcal{F}^{(s)}_{\mathrm{grp}}(n^0,H^{(s)})_i$.
Fusing all branches gives the agent features $F_i=\mathrm{LN}\big(W_o[n^0_i;\,n^{\mathrm{pair}}_i;\,n^{(s_1)}_i;\dots;\,n^{(s_{|\mathcal{S}|})}_i]\big)$, and the final hyperedge features give the encoder group tokens $\Gamma=\mathrm{LN}(W_\gamma[h_e]_{e=1}^{E})$.

\subsubsection{Hypothesis-Conditioned Decoder}
\label{sec:hypgroups}
\noindent\textbf{Query Tokens and Self-conditioning.}
Each decoder token $G^{(0)}_{k,i}=\mathrm{MLP}_2\big(\mathrm{MLP}_1([F_i;\,U_{k,i};\,\phi_t])+\omega_k+\nu_i\big)$ fuses the agent feature $F_i$, an embedding $U_{k,i}$ of the path point $Y^t_{k,i}$ scaled to unit variance, the flow-time embedding $\phi_t$, and learned embeddings $\omega_k$ of the hypothesis and $\nu_i$ of the agent.
For self-conditioning~\cite{chen2023analogbits}, $U_{k,i}$ also adds an embedding of the estimate $\hat Y_{k,i}$, the previous step's endpoint.

\noindent\textbf{Futures Define Groups.}
Each hypothesis groups agents by the similarity of their tokens and their proximity in its own estimated future.
For each of $M=2$ horizon segments with frames $\mathcal{T}_m$, block $b$ measures the mean distance of agents $i$ and $j$ in the estimated future of hypothesis $k$,
\begin{equation}
  \delta^{m}_{k,ij}=\frac{M}{T_f}\sum_{\tau\in\mathcal{T}_m}\big\lVert (x_i+\hat Y_{k,i,\tau})-(x_j+\hat Y_{k,j,\tau})\big\rVert_2,
\end{equation}
and subtracts it from the token similarity,
\begin{equation}
  \alpha^{b,m}_{k,ij}=\bar g_{k,i}^{\top}\bar g_{k,j}-t\,w\,\delta^{m}_{k,ij},
  \label{eq:futaff}
\end{equation}
where $\bar g_{k,i}$ is the centered and normalized projection of $G_{k,i}$ and $w>0$ is learned.
As the future term grows with $t$, the groups shift from token similarity toward proximity in the estimated future as it forms.
The hyperedge builder (\cref{sec:encoder}) with this affinity forms the groups of hypothesis $k$ for segment $m$ in block $b$,
\begin{equation}
  H^{b,m}_k=\mathrm{Group}^{b,m}\big(G_{k,:},\,\hat Y_k,\,t\big),
  \label{eq:hcg}
\end{equation}
so every hypothesis has its own hyperedges of several sizes, rebuilt in every block and at every flow step.

\noindent\textbf{Groups Shape Futures.}
In turn, the groups shape the future of their hypothesis through group tokens, a group-factorized endpoint, and the hyperedge-level loss of \cref{sec:train_sample}.
Every group becomes a token, the mean of its members' tokens plus a segment embedding $\psi_m$,
\begin{equation}
  \gamma^{b,m}_{k,e}=\mathrm{LN}\Big(W^{b}_h\frac{1}{|e|}\sum_iH^{b,m}_{k,i,e}G_{k,i}\Big)+\psi_m,
\end{equation}
and in the agent attention of block $b$, the agent tokens of hypothesis $k$ attend to each other, to the encoder group tokens, and to its own $ME$ group tokens,
\begin{equation}
  G_{k,:}\leftarrow\mathrm{AgentAttn}_b\big([G_{k,:};\,\Gamma;\,\gamma^{b,1}_{k};\,\gamma^{b,2}_{k}]\big)_{1:N},
  \label{eq:block}
\end{equation}
keeping only the $N$ agent outputs.
\label{sec:gf}
The group-factorized endpoint couples the members of each group through a shared correction: $S_{k,i}$ adds to a per-agent term the mean correction that the groups containing agent $i$ predict from their tokens in the last block $B$,
\begin{equation}
  S_{k,i}=\mathrm{MLP}_{\mathrm{reg}}\big(G^{(B)}_{k,i}\big)+\frac{\sum_{m,e}\tilde H^{B,m}_{k,i,e}\,\mathrm{MLP}_{\mathrm{grp}}\big(\gamma^{B,m}_{k,e}\big)}{\sum_{m,e}\tilde H^{B,m}_{k,i,e}}.
  \label{eq:gf}
\end{equation}
The straight-through incidence $\tilde H$ lets the forecasting loss train the hyperedge builder.
The logits are $\zeta_{k,i}=\mathrm{MLP}_{\mathrm{cls}}(G^{(B)}_{k,i})$, and \cref{eq:vel} gives the velocity of each flow.

\subsection{Training}
\label{sec:train_sample}
Training samples $t\sim\mathrm{LogitNormal}(-0.5,1.5)$ and one noise draw $Y^0$ per scene (\cref{sec:bg_moflow}), and minimizes
\begin{equation}
  \mathcal{L}=\mathbb{E}_{t,Y^0,Y}\big[\mathcal{L}_{\mathrm{agent}}+\mathcal{L}_{\mathrm{cls}}+\lambda_q\,\mathcal{L}_{\mathrm{grp}}\big],
  \label{eq:total}
\end{equation}
with $\mathcal{L}_{\mathrm{agent}}$ and $\mathcal{L}_{\mathrm{cls}}$ of \cref{sec:bg_moflow}, $d_{k,i}$ in meters, and a hyperedge-level loss $\mathcal{L}_{\mathrm{grp}}$ with weight $\lambda_q$ at epoch $q$.

\noindent\textbf{Hyperedge-level Winner-take-all.}
$\mathcal{L}_{\mathrm{grp}}$ scores each hypothesis on its own groups and keeps the best hypothesis per hyperedge,
\begin{equation}
  \mathcal{L}_{\mathrm{grp}}=\frac{1}{ME}\sum_{e=1}^{ME}\min_k\,\epsilon_{k,e}.
  \label{eq:grp}
\end{equation}
An index $e$ fixes an anchor, a size, and a segment; $e_k$ is the group of hypothesis $k$ at this index in the last block, with incidence $H_{k,:,e}=H^{B,m}_{k,:,e}$, and $\epsilon_{k,e}=\sum_iH_{k,i,e}\,d_{k,i}/|e_k|$ is its mean error.
A hypothesis wins a hyperedge only by predicting its whole group well, and different hyperedges can select different winners, so the loosely coupled groups of one scene may follow different hypotheses.
The incidence is detached, so $\mathcal{L}_{\mathrm{grp}}$ cannot favor easy-to-predict groups.

\noindent\textbf{Loss Schedule and Self-conditioning.}
As $\mathcal{L}_{\mathrm{grp}}$ trades per-agent for joint accuracy (\cref{sec:ablation}), its weight decays over the $J$ epochs as $\lambda_q=\frac{\lambda_0}{2}\big(1+\cos(\pi q/J)\big)$, with $\lambda_0=0.25$.
For self-conditioning, half the batches first estimate $\hat Y$ without gradients and condition on it; the others use none.

\subsection{Inference}
\label{sec:inference}
One noise sample $Y^0$ starts all $K$ flows, which thus differ at the first step only through $\omega_k$.
At every Euler step $t\in\{0,\Delta t,\dots,1-\Delta t\}$, the decoder maps $(Y^t,t,\hat Y)$ to the endpoints $S_k$, logits $\zeta_k$, and groups $H_k$, and each flow moves along its own velocity, $Y^{t+\Delta t}_k=Y^t_k+\Delta t\,(S_k-Y^t_k)/(1-t)$.
The endpoints of one step are the estimate $\hat Y$ of the next, so every hypothesis rebuilds its groups from its latest estimate (\cref{eq:futaff}) at no extra network evaluation.
The last step reaches $Y^1_k=S_k$, so agent $i$ receives the $K$ trajectories $x_i+S_{k,i}$, scored by $\zeta_{k,i}$.

\section{Experiments}
\label{sec:experiments}

\begin{table*}[!t]
  \centering
  \caption{Results on team sports at 4.0s: min$_{20}$ADE/FDE (m) per agent, per three-nearest-neighbor group (G3), per team, and per scene, with all models retrained on the same scenes. Improvement: over the best other method. Best in bold, second underlined.}
  \label{tab:sports}
  \footnotesize
  \setlength{\tabcolsep}{2.5pt}
  \begin{adjustbox}{max width=\textwidth}
  \begin{tabular}{lccccccccccccc}
    \toprule
    & & \multicolumn{4}{c}{NBA} & \multicolumn{4}{c}{NFL} & \multicolumn{4}{c}{Soccer} \\
    \cmidrule(lr){3-6} \cmidrule(lr){7-10} \cmidrule(l){11-14}
    Method & Venue & Agent & G3 & Team & Scene & Agent & G3 & Team & Scene & Agent & G3 & Team & Scene \\
    \midrule
    GroupNet~\cite{xu2022groupnet} & \venue{CVPR'22} & 0.891/1.224 & 1.512/2.610 & 1.632/3.071 & 2.043/3.798 & 1.918/3.301 & 3.390/6.610 & 4.586/9.299 & 5.174/10.508 & 1.104/2.016 & 1.945/4.045 & 2.354/5.160 & 2.821/6.079 \\ %
    LED~\cite{mao2023led} & \venue{CVPR'23} & 0.808/1.152 & 1.346/2.267 & 1.404/2.579 & 1.593/2.854 & 1.470/2.649 & 2.399/4.615 & 2.916/5.849 & 3.159/6.315 & 1.061/1.890 & 1.675/3.233 & 1.676/3.390 & 1.921/3.834 \\ %
    MART~\cite{lee2024mart} & \venue{ECCV'24} & 0.736/0.924 & 1.281/2.150 & 1.367/2.546 & 1.540/2.792 & 1.209/1.943 & 2.074/3.940 & 2.570/5.164 & 2.773/5.556 & 1.003/1.757 & 1.584/3.037 & \textbf{1.584}/\underline{3.199} & 1.824/3.618 \\ %
    MoFlow~\cite{fu2025moflow} & \venue{CVPR'25} & \underline{0.699}/\underline{0.878} & \underline{1.219}/\underline{2.067} & \underline{1.310}/\underline{2.485} & \underline{1.463}/\underline{2.695} & \underline{1.104}/\underline{1.794} & \underline{1.887}/\underline{3.649} & \underline{2.358}/\underline{4.816} & \underline{2.499}/\underline{5.086} & 1.091/1.766 & 1.727/3.077 & 1.798/3.359 & 2.028/3.730 \\
    Social-Mamba~\cite{luan2026socialmamba} & \venue{ECCV'26} & 0.729/0.925 & 1.284/2.175 & 1.368/2.585 & 1.534/2.814 & 1.168/1.922 & 2.020/3.909 & 2.533/5.193 & 2.704/5.499 & \textbf{0.938}/\underline{1.637} & \underline{1.556}/\underline{3.017} & 1.590/3.254 & \textbf{1.797}/\underline{3.615} \\ %
    \midrule
    \oursrow H\textsuperscript{2}Flow & \venue{Ours} & \textbf{0.673}/\textbf{0.842} & \textbf{1.143}/\textbf{1.959} & \textbf{1.227}/\textbf{2.346} & \textbf{1.372}/\textbf{2.554} & \textbf{1.081}/\textbf{1.743} & \textbf{1.811}/\textbf{3.511} & \textbf{2.274}/\textbf{4.669} & \textbf{2.405}/\textbf{4.922} & \underline{0.964}/\textbf{1.611} & \textbf{1.534}/\textbf{2.864} & \underline{1.588}/\textbf{3.132} & \underline{1.799}/\textbf{3.487} \\
    Improvement & & 3.7\%/4.1\% & 6.2\%/5.2\% & 6.3\%/5.6\% & 6.2\%/5.2\% & 2.1\%/2.8\% & 4.0\%/3.8\% & 3.5\%/3.1\% & 3.8\%/3.2\% & $-$2.7\%/1.6\% & 1.4\%/5.1\% & $-$0.3\%/2.1\% & $-$0.1\%/3.5\% \\ %
    \bottomrule
  \end{tabular}
  \end{adjustbox}
\end{table*}

\begin{table}[t]
  \centering
  \caption{Per-agent min$_{20}$ADE/FDE (m) on NBA at 1.0--4.0s. Best in bold, second underlined.}
  \label{tab:nba}
  \footnotesize
  \setlength{\tabcolsep}{6pt}
  \begin{adjustbox}{max width=\columnwidth}
  \begin{tabular}{lccccc}
    \toprule
    Method & Venue & 1.0s & 2.0s & 3.0s & 4.0s \\
    \midrule
    MemoNet~\cite{xu2022memonet} & \venue{CVPR'22} & 0.38/0.56 & 0.71/1.14 & 1.00/1.57 & 1.25/1.47 \\
    NPSN~\cite{bae2022npsn} & \venue{CVPR'22} & 0.35/0.58 & 0.68/1.23 & 1.01/1.76 & 1.31/1.79 \\
    GroupNet~\cite{xu2022groupnet} & \venue{CVPR'22} & 0.21/0.32 & 0.42/0.67 & 0.66/0.98 & 0.89/1.22 \\ %
    MID~\cite{gu2022mid} & \venue{CVPR'22} & 0.28/0.37 & 0.51/0.72 & 0.71/0.98 & 0.96/1.27 \\
    LED-D~\cite{mao2023led} & \venue{CVPR'23} & 0.21/0.28 & 0.44/0.64 & 0.69/0.95 & 0.94/1.21 \\
    LED-I~\cite{mao2023led} & \venue{CVPR'23} & \underline{0.18}/0.27 & 0.37/0.56 & 0.58/0.83 & 0.81/1.15 \\ %
    DynGroupNet~\cite{xu2022dyngroupnet} & \venue{NN'24} & 0.19/0.28 & 0.40/0.61 & 0.65/0.90 & 0.89/1.13 \\
    MART~\cite{lee2024mart} & \venue{ECCV'24} & \underline{0.18}/\underline{0.26} & 0.36/0.51 & 0.54/0.72 & 0.74/0.92 \\ %
    MoFlow~\cite{fu2025moflow} & \venue{CVPR'25} & \underline{0.18}/\textbf{0.25} & \textbf{0.33}/\underline{0.46} & \underline{0.51}/\underline{0.67} & \underline{0.70}/\underline{0.88} \\ %
    Social-Mamba~\cite{luan2026socialmamba} & \venue{ECCV'26} & \textbf{0.17}/\textbf{0.25} & \underline{0.34}/0.50 & 0.53/0.72 & 0.73/0.92 \\ %
    \midrule
    \oursrow H\textsuperscript{2}Flow & \venue{Ours} & \textbf{0.17}/\textbf{0.25} & \textbf{0.33}/\textbf{0.45} & \textbf{0.49}/\textbf{0.64} & \textbf{0.67}/\textbf{0.84} \\
    \bottomrule
  \end{tabular}
  \end{adjustbox}
\end{table}

\subsection{Datasets and Metrics}
\label{sec:data}

\noindent\textbf{Datasets.}
NBA SportVU (10 players and the ball) follows the standard protocol~\cite{mao2023led,fu2025moflow}: 2.0s of past (10 frames) predict 4.0s (20 frames) at 5Hz, with 32.5k/12.5k training/test scenes.
NFL (2017 season) and soccer (SoccerTrack~\cite{scott2022soccertrack}, one match), with 22 players and the ball, use the same frames and the splits of Sports-Traj~\cite{xu2025sportstraj} (23.1k/3.5k and 32.2k/684 scenes).
ETH-UCY~\cite{pellegrini2009eth,lerner2007ucy} (SocialGAN split~\cite{gupta2018socialgan}) and SDD~\cite{robicquet2016sdd} (TUTR split~\cite{shi2023tutr}) predict 4.8s (12 frames) from 3.2s (8)~\cite{fu2025moflow}, on the baselines' trajectories.

\noindent\textbf{Metrics.}
All metrics use $K=20$ hypotheses, in meters (pixels on SDD).
min$_{20}$ADE/FDE match each agent to its best hypothesis; G3-, Team-, and Scene-min$_{20}$ADE/FDE match each group (the 3 nearest neighbors of each agent at the last observed frame, a team, or all agents) to the hypothesis with the lowest mean error over its members and average over groups, so Scene equals JADE/JFDE~\cite{weng2023joint}.
H\textsuperscript{2}Flow trains on none of these groupings, so G3, Team, and Scene measure joint accuracy at three scales independently of the learned groups.
We evaluate per scene, whereas the public code of MoFlow selects one hypothesis index per batch of 1,000 test scenes.

\subsection{Implementation Details}
\label{sec:impl}
We build on the official MoFlow code, whose decoder blocks all read the initial tokens $G^{(0)}$, so only the last block is effective; we chain the blocks (\cref{eq:block}).
MoFlow is retrained with this decoder and our training recipe on NFL and soccer and in all team-sport experiments beyond \cref{tab:sports,tab:nba}, and with its released code otherwise.
H\textsuperscript{2}Flow uses $D=128$, $L=4$ encoder layers, group sizes $\mathcal{S}=\{2,3,5\}$, $B=4$ decoder blocks, and $M=2$ segments of 2.0s.
On NBA, we train with AdamW and a cosine schedule with warm-up for $J=100$ epochs, also in the ablations, and MoFlow for 150 epochs; sampling uses 10 Euler steps.
Results average three runs of H\textsuperscript{2}Flow and two of MoFlow on NBA, two of each on NFL and soccer, and three of each on ETH-UCY and SDD; other entries use one run (settings in the \suppapp{supplement}{appendix}).

\subsection{Baselines}
\label{sec:baselines}
On NBA, we compare with the baselines reported by MoFlow~\cite{fu2025moflow} (MemoNet~\cite{xu2022memonet}, NPSN~\cite{bae2022npsn}, MID~\cite{gu2022mid}, LED~\cite{mao2023led}, and GroupNet~\cite{xu2022groupnet}), the group-aware DynGroupNet~\cite{xu2022dyngroupnet} and MART~\cite{lee2024mart}, the state-space Social-Mamba~\cite{luan2026socialmamba}, and MoFlow itself.
For \cref{tab:sports}, LED, GroupNet, MART, Social-Mamba, MoFlow, and H\textsuperscript{2}Flow are trained on the same scenes of all three sports.

\subsection{Quantitative Results}
\label{sec:quant}

\noindent\textbf{NBA.}
H\textsuperscript{2}Flow is the most accurate per agent from 2.0s to 4.0s (\cref{tab:nba}); at 4.0s, its min$_{20}$ADE/FDE of 0.67/0.84 is 3.7\%/4.1\% below MoFlow (0.699/0.878, matching the published 0.70/0.88).
The gains grow where several agents must be right together (\cref{tab:sports}): MoFlow's best hypothesis per scene errs twice as much as its best per agent (1.463 \vs 0.699), and H\textsuperscript{2}Flow lowers G3-, Team-, and Scene-min$_{20}$ADE by 6.2--6.3\%, beyond its run spread (1.368--1.375).
The scene encoder alone already matches it per agent (0.672/0.841, three runs of 150 epochs), so the hypothesis-conditioned groups and $\mathcal{L}_{\mathrm{grp}}$ lower the group-level errors by a further 2.8--3.4\%.

\noindent\textbf{NFL and Soccer.}
The gains carry over to 22 players and the ball (\cref{tab:sports}): H\textsuperscript{2}Flow lowers min$_{20}$ADE by 2.1\% on NFL and 11.7\% on soccer and Scene-min$_{20}$ADE by 3.8\% and 11.3\% relative to MoFlow, beyond the run spread.

\begin{table}[t]
  \centering
  \caption{Pedestrian results at 4.8s: min$_{20}$ADE/FDE on ETH-UCY (m) and SDD (pixels). Best in bold, second underlined.}
  \label{tab:ped}
  \footnotesize
  \setlength{\tabcolsep}{2pt}
  \begin{adjustbox}{max width=\columnwidth}
  \begin{tabular}{lcccccccc}
    \toprule
    & & \multicolumn{6}{c}{ETH-UCY} & \\
    \cmidrule(lr){3-8}
    Method & Venue & ETH & Hotel & Univ & Zara1 & Zara2 & AVG & SDD \\
    \midrule
    MID~\cite{gu2022mid} & \venue{CVPR'22} & .39/.66 & .13/.22 & \textbf{.22}/.45 & \underline{.17}/.30 & \underline{.13}/.27 & \textbf{.21}/.38 & 9.73/15.32 \\
    GroupNet~\cite{xu2022groupnet} & \venue{CVPR'22} & .46/.73 & .15/.25 & .26/.49 & .21/.39 & .17/.33 & .25/.44 & 9.31/16.11 \\
    LED~\cite{mao2023led} & \venue{CVPR'23} & .39/.58 & \textbf{.11}/\underline{.17} & .26/.44 & .18/\textbf{.26} & \underline{.13}/\textbf{.22} & \textbf{.21}/\underline{.33} & 8.48/\textbf{11.66} \\
    TUTR~\cite{shi2023tutr} & \venue{ICCV'23} & .40/.61 & \textbf{.11}/.18 & \underline{.23}/.42 & .18/.34 & \underline{.13}/.25 & \textbf{.21}/.36 & 7.76/12.69 \\
    EigenTraj~\cite{bae2023eigentrajectory} & \venue{ICCV'23} & \underline{.36}/\underline{.53} & \underline{.12}/.19 & .24/\textbf{.34} & .19/.33 & .14/.24 & \textbf{.21}/\underline{.33} & 8.05/13.25 \\
    SingularTraj~\cite{bae2024singulartrajectory} & \venue{CVPR'24} & \textbf{.35}/\textbf{.42} & .13/.19 & .25/.44 & .19/.32 & .15/.25 & \textbf{.21}/\textbf{.32} & 8.40/13.99 \\ %
    MoFlow~\cite{fu2025moflow} & \venue{CVPR'25} & .42/.59 & \underline{.12}/.18 & \underline{.23}/\underline{.40} & \underline{.17}/\underline{.29} & \textbf{.12}/\textbf{.22} & \textbf{.21}/.34 & \underline{7.68}/12.33 \\ %
    Social-Mamba~\cite{luan2026socialmamba} & \venue{ECCV'26} & .43/.79 & .19/.40 & .32/.60 & .18/\underline{.29} & .14/\underline{.23} & .25/.46 & 8.16/13.46 \\ %
    \midrule
    \oursrow H\textsuperscript{2}Flow & \venue{Ours} & .44/.64 & \underline{.12}/\textbf{.16} & \underline{.23}/.41 & \textbf{.16}/\underline{.29} & \underline{.13}/\textbf{.22} & \underline{.22}/.34 & \textbf{7.44}/\underline{11.92} \\ %
    \bottomrule
  \end{tabular}
  \end{adjustbox}
\end{table}

\noindent\textbf{Pedestrian Benchmarks.}
On ETH-UCY and SDD, H\textsuperscript{2}Flow has the lowest min$_{20}$ADE on Zara1 and SDD and the lowest min$_{20}$FDE on Hotel (tied on Zara2), but trails most methods on ETH (\cref{tab:ped}).
Its best configurations forecast each pedestrian alone on ETH, Univ, Zara2, and SDD, so these benchmarks test its decoder more than its groups.

\subsection{Qualitative Results}
\label{sec:qual}
\begin{figure*}[t]
  \centering
  \includegraphics[width=\appscale{0.97}\textwidth]{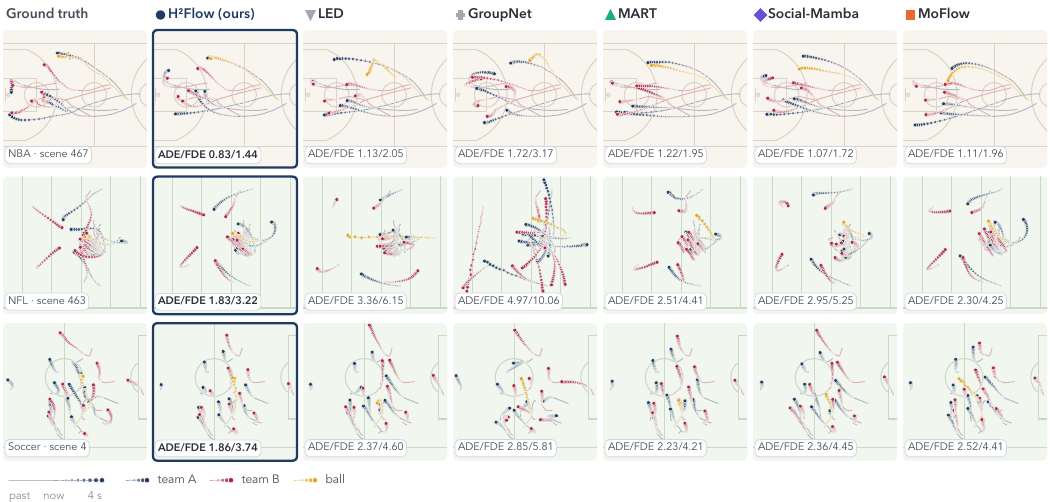}  %
  \caption{Best joint hypothesis of each model on an NBA, an NFL, and a soccer play, with its ADE/FDE (m) in the badge.}
  \label{fig:qual_pred_gt}
\end{figure*}

\begin{figure*}[t]
  \centering
  \includegraphics[width=\appscale{0.97}\textwidth]{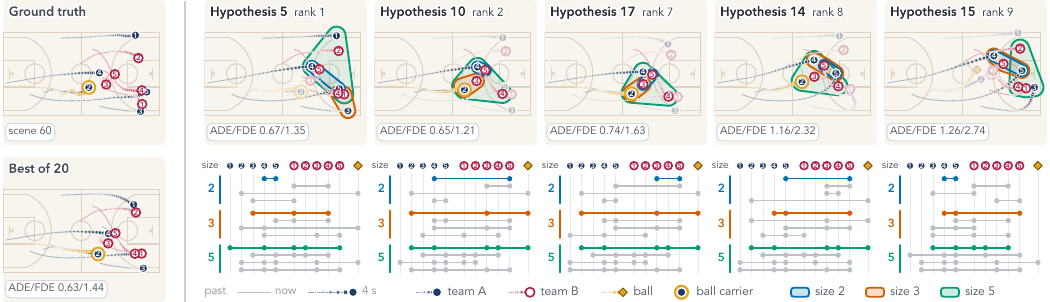}  %
  \caption{Hypotheses of H\textsuperscript{2}Flow on one NBA play with their score rank and the most-supported group of each size (top), and their four best-supported hyperedges per size over 2.0--4.0s (bottom).}
  \label{fig:qual_hyp_groups}
\end{figure*}

On an NBA, an NFL, and a soccer play (\cref{fig:qual_pred_gt}), the best joint hypothesis of H\textsuperscript{2}Flow follows the true future most closely (ADE over all agents 0.83, 1.83, and 1.86m, against at least 1.07, 2.30, and 2.23m for the baselines).
The hypotheses of one play (\cref{fig:qual_hyp_groups}) form different groups, each from its own estimated future; 10 of the 15 shown are distinct, so the hypotheses differ in which players move together as well as in where they go.
Fixed rules select and rank the candidate plays; \suppapp{the supplementary material}{the appendix} adds more plays, all 20 futures of two agents, and the groups over the flow steps.

\subsection{Ablation Studies}
\label{sec:ablation}
\begin{table}[t]
  \centering
  \caption{Ablation on NBA at 4.0s (m; abbreviations in \cref{sec:ablation}). Best in bold, second underlined.}
  \label{tab:ablation}
  \footnotesize
  \setlength{\tabcolsep}{2pt}
  \begin{adjustbox}{max width=\columnwidth}
  \begin{tabular}{@{}lccccccc@{}}
    \toprule
    \multicolumn{4}{@{}l}{\emph{(a) Groups shape futures}} & Per agent & \multicolumn{3}{c}{Per group (best hypothesis)} \\
    \cmidrule(lr){5-5}\cmidrule(l){6-8}
    Variant & GT & GF & $\mathcal{L}_{\mathrm{grp}}$ & min$_{20}$ADE/FDE$\downarrow$ & G3$\downarrow$ & Team$\downarrow$ & Scene$\downarrow$ \\
    \midrule
    \oursrow H\textsuperscript{2}Flow & \checkmark & \checkmark & \checkmark & \textbf{0.672}/\textbf{0.840} & \underline{1.144} & \underline{1.226} & \underline{1.373} \\
    w/o GT &  & \checkmark & \checkmark & 0.678/0.854 & 1.148 & 1.233 & 1.377 \\
    w/o GF & \checkmark &  & \checkmark & 0.679/0.858 & \textbf{1.139} & \textbf{1.223} & \textbf{1.365} \\
    w/o $\mathcal{L}_{\mathrm{grp}}$ & \checkmark & \checkmark &  & 0.692/0.871 & 1.227 & 1.307 & 1.464 \\
    w/o GT, GF &  &  & \checkmark & 0.681/0.854 & \underline{1.144} & \underline{1.226} & \underline{1.373} \\
    w/o GT, $\mathcal{L}_{\mathrm{grp}}$ &  & \checkmark &  & 0.690/0.869 & 1.226 & 1.309 & 1.466 \\
    w/o GF, $\mathcal{L}_{\mathrm{grp}}$ & \checkmark &  &  & \underline{0.675}/\underline{0.848} & 1.189 & 1.282 & 1.437 \\
    \midrule
    \multicolumn{4}{@{}l}{\emph{(b) Futures define groups}} & & & & \\
    Variant & SC & FUT & TW & & & & \\
    \midrule
    \oursrow H\textsuperscript{2}Flow & \checkmark & \checkmark & \checkmark & \textbf{0.672}/\underline{0.840} & 1.144 & \underline{1.226} & 1.373 \\
    w/o TW & \checkmark & \checkmark &  & \textbf{0.672}/\textbf{0.839} & 1.145 & 1.230 & 1.375 \\
    w/o FUT & \checkmark &  &  & \underline{0.673}/0.849 & \textbf{1.141} & \underline{1.226} & \underline{1.372} \\
    w/o SC &  &  &  & 0.677/0.850 & \underline{1.143} & \textbf{1.223} & \textbf{1.366} \\
    \bottomrule
  \end{tabular}
  \end{adjustbox}
\end{table}
\begin{table}[t]
  \centering
  \caption{Group loss $\mathcal{L}_{\mathrm{grp}}$ on NBA at 4.0s (m): (a) scored groups (random and true-future ones of the hypothesis-group sizes); (b) initial weight $\lambda_0$ and schedule. Best in bold, second underlined.}
  \label{tab:loss}
  \footnotesize
  \setlength{\tabcolsep}{3pt}
  \begin{adjustbox}{max width=\columnwidth}
  \begin{tabular}{lccccc}
    \toprule
    & Per agent & \multicolumn{3}{c}{Per group (best hypothesis)} & Final epoch \\
    \cmidrule(lr){2-2}\cmidrule(lr){3-5}\cmidrule(lr){6-6}
    Variant & min$_{20}$ADE/FDE$\downarrow$ & G3$\downarrow$ & Team$\downarrow$ & Scene$\downarrow$ & min$_{20}$ADE$\downarrow$ \\
    \midrule
    \multicolumn{6}{l}{\emph{(a) Groups scored by $\mathcal{L}_{\mathrm{grp}}$ ($\lambda_0=0.25$, cosine decay)}} \\
    \rowcolor{black!6}none (no $\mathcal{L}_{\mathrm{grp}}$) & 0.692/0.871 & 1.227 & 1.307 & 1.464 & 0.692 \\
    \oursrow hypothesis groups (H\textsuperscript{2}Flow) & \textbf{0.672}/\textbf{0.840} & 1.144 & 1.226 & 1.373 & \textbf{0.679} \\
    encoder groups & \underline{0.674}/0.846 & 1.140 & \underline{1.213} & 1.365 & \underline{0.685} \\
    random groups & 0.676/0.851 & \textbf{1.129} & \textbf{1.208} & \underline{1.352} & 0.689 \\
    true-future groups & \textbf{0.672}/\underline{0.844} & 1.151 & 1.255 & 1.400 & \textbf{0.679} \\
    whole scene & 0.676/0.845 & \underline{1.137} & 1.214 & \textbf{1.345} & 0.690 \\
    \cmidrule{1-6}
    \emph{average of the five} & 0.674/0.845 & 1.140 & 1.223 & 1.367 & 0.684 \\
    \midrule
    \multicolumn{6}{l}{\emph{(b) Weight $\lambda_0$ and schedule of $\mathcal{L}_{\mathrm{grp}}$ on the hypothesis groups}} \\
    \oursrow $\lambda_0=0.25$, cosine (H\textsuperscript{2}Flow) & \textbf{0.672}/\textbf{0.840} & 1.144 & 1.226 & 1.373 & \underline{0.679} \\
    $\lambda_0=0.1$, cosine & \underline{0.676}/\underline{0.847} & 1.178 & 1.261 & 1.411 & \textbf{0.676} \\
    $\lambda_0=0.5$, cosine & 0.684/0.865 & \textbf{1.128} & \underline{1.211} & \textbf{1.353} & 0.707 \\
    $\lambda_0=1$, cosine & 0.707/0.901 & \textbf{1.128} & \textbf{1.205} & \underline{1.356} & 0.796 \\
    $\lambda_0=0.25$, constant & 0.677/0.852 & \underline{1.131} & 1.218 & 1.361 & 0.695 \\
    \bottomrule
  \end{tabular}
  \end{adjustbox}
\end{table}

\cref{tab:ablation} varies two sets of parts: (a) the hypothesis group tokens (GT), the group-factorized endpoint (GF), and $\mathcal{L}_{\mathrm{grp}}$, through which the groups shape the futures, removed alone and in pairs; (b) self-conditioning (SC), the future term (FUT), and its time weight (TW), through which the futures define the groups, each removed with its dependent parts.
\cref{tab:loss} varies the groups $\mathcal{L}_{\mathrm{grp}}$ scores (a) and its weight (b).

\noindent\textbf{Groups Shape Futures.}
$\mathcal{L}_{\mathrm{grp}}$ drives joint accuracy (\cref{tab:ablation}a): removing it from H\textsuperscript{2}Flow or from the variants without GT or without GF raises Scene-min$_{20}$ADE by 0.072--0.091, and without it the endpoint raises the error further (1.464 \vs 1.437).
Alone, $\mathcal{L}_{\mathrm{grp}}$ matches the Scene-min$_{20}$ADE of H\textsuperscript{2}Flow (1.373) at a higher min$_{20}$ADE/FDE (0.681/0.854); the tokens and the endpoint lower it to 0.672/0.840 only together (0.679/0.858 with the tokens alone, 0.678/0.854 with the endpoint alone), while removing either moves the group-level errors by at most 0.008.

\noindent\textbf{Futures Define Groups.}
Forming the groups from the estimated futures mainly affects the per-agent error (\cref{tab:ablation}b): removing TW or FUT moves min$_{20}$ADE by at most 0.001 and the group-level errors by at most 0.004, and removing SC as well raises min$_{20}$ADE/FDE to 0.677/0.850 while the group-level errors stay within 0.007.

\noindent\textbf{Scoring the Hypotheses on Groups.}
Every group source lets $\mathcal{L}_{\mathrm{grp}}$ lower Scene-min$_{20}$ADE by 0.064--0.119 (\cref{tab:loss}a), so on NBA its gain comes from selecting one winner per hyperedge; over three runs, encoder groups even reach 1.366 (\vs 1.372 for H\textsuperscript{2}Flow).
The sources trade joint for per-agent accuracy: random groups and the whole scene give the lowest G3- and Scene-min$_{20}$ADE (1.352 and 1.345 for Scene), and the hypothesis and true-future groups the lowest per-agent error and the least drift (0.672, final 0.679).
The weight sets the same trade-off (\cref{tab:loss}b); cosine decay limits a constant weight's drift (final 0.679 \vs 0.695).

\subsection{Hypothesis Group Analysis}
\label{sec:analysis}
\begin{table}[t]
  \centering
  \caption{Hypothesis groups on NBA: (a) the groups formed for 2.0--4.0s; (b) errors at 4.0s (m) with other groups read at test time. H\textsuperscript{2}Flow and (b): three runs. Best in bold, second underlined.}
  \label{tab:groups}
  \footnotesize
  \begin{adjustbox}{max width=\columnwidth}
  \vbox{\halign{\hfil#\hfil\cr
  \setlength{\tabcolsep}{2.5pt}
  \begin{tabular}{@{}lcccccc@{}}
    \toprule
    \multicolumn{1}{@{}l}{\emph{(a) Groups formed}} & \multicolumn{2}{c}{Groups differ (\%)} & \multicolumn{3}{c}{Overlap with the groups of} & Own future \\
    \cmidrule(lr){2-3}\cmidrule(lr){4-6}\cmidrule(l){7-7}
    Variant & \shortstack{between\\hypotheses} & \shortstack{between\\flow steps} & \shortstack{own\\future$\uparrow$} & \shortstack{another\\hyp.'s future} & \shortstack{true\\future$\uparrow$} & \shortstack{closest\\($\times$chance)$\uparrow$} \\
    \midrule
    \oursrow H\textsuperscript{2}Flow & 38.8 & 3.9 & \underline{0.687} & 0.654 & \textbf{0.581} & \textbf{2.6}$\times$ \\
    w/o $\mathcal{L}_{\mathrm{grp}}$ & 35.2 & 3.7 & \textbf{0.704} & 0.676 & \underline{0.576} & \underline{2.5}$\times$ \\
    w/o GF, $\mathcal{L}_{\mathrm{grp}}$ & 44.8 & 3.5 & 0.623 & 0.604 & 0.546 & 1.9$\times$ \\
    \hspace{0.8em}+ w/o SEG, TW & 45.7 & 3.5 & 0.616 & 0.610 & 0.545 & 1.3$\times$ \\
    \hspace{0.8em}+ w/o FUT & 51.2 & 0.9 & 0.539 & 0.541 & 0.499 & 0.9$\times$ \\
    \hspace{0.8em}+ w/o SC & 55.4 & 0.7 & 0.534 & 0.542 & 0.498 & 0.8$\times$ \\
    \bottomrule
  \end{tabular}\cr\noalign{\vskip 4pt}
  \setlength{\tabcolsep}{3pt}
  \begin{tabular}{@{}lcccc@{}}
    \toprule
    \multicolumn{1}{@{}l}{\emph{(b) At test time, H\textsuperscript{2}Flow reads}} & Per agent & \multicolumn{3}{c}{Per group (best hypothesis)} \\
    \cmidrule(lr){2-2}\cmidrule(l){3-5}
    Groups & min$_{20}$ADE$\downarrow$ & G3$\downarrow$ & Team$\downarrow$ & Scene$\downarrow$ \\
    \midrule
    \oursrow own groups (H\textsuperscript{2}Flow) & \textbf{0.6728} & \textbf{1.1429} & \textbf{1.2271} & \textbf{1.3720} \\
    groups of another hypothesis & 0.6732 & 1.1432 & 1.2275 & 1.3726 \\
    groups of one hypothesis, for all & 0.6731 & 1.1435 & 1.2279 & 1.3728 \\
    groups frozen after the first flow step & \underline{0.6729} & \underline{1.1431} & \underline{1.2272} & \underline{1.3722} \\
    nearest-neighbor groups of its own estimate & 0.6735 & 1.1437 & 1.2275 & 1.3725 \\
    random groups of the same sizes & 0.6766 & 1.1488 & 1.2307 & 1.3770 \\
    encoder groups & 0.6773 & 1.1500 & 1.2312 & 1.3775 \\
    \bottomrule
  \end{tabular}\cr}}
  \end{adjustbox}
\end{table}

\cref{tab:groups}a scores the 2.0--4.0s groups of the last decoder block (SEG: the two segments of the future term) by their Jaccard overlap with the nearest-neighbor groups of three futures and by how often its own future is the closest of the 20; \cref{tab:groups}b swaps the groups the decoder reads at test time.

\noindent\textbf{Every Hypothesis Forms the Groups of Its Own Future.}
A hypothesis's groups overlap most with those of its own future (0.687, \vs 0.654 for another hypothesis's, 0.581 for the true future, and 0.21 for random groups), which is the closest of the 20 at 2.6$\times$ chance.
Groups from the tokens alone differ more between hypotheses (51.2--55.4\% \vs 38.8\%) but match their own future at chance (0.8--0.9$\times$); the future term, its segments, and the endpoint raise the match to 1.3$\times$, 1.9$\times$, and 2.5$\times$.
77\% of hyperedges join players of both teams.

\noindent\textbf{The Decoder Reads Groups Formed from Predicted Motion.}
Random or encoder groups raise Scene-min$_{20}$ADE by 0.50 and 0.55\,cm, whereas the other swaps of \cref{tab:groups}b change it by only 0.02--0.08\,cm: the decoder depends on groups formed from predicted motion, and only slightly on which hypothesis formed them.

\noindent\textbf{One Grouping per Play.}
Trained with one grouping per play, formed from the mean tokens and estimated distances of its hypotheses, the decoder is as accurate as H\textsuperscript{2}Flow (min$_{20}$ADE and Scene-min$_{20}$ADE 0.13 and 0.11\,cm higher over three runs; differences between paired runs $-$0.7 to $+$0.9\,cm).
On NBA, the accuracy thus comes from forming groups from the estimated futures and scoring the hypotheses on them; per-hypothesis groups add each hypothesis's match with its own future at no extra network evaluation.

\noindent\textbf{Groups and Joint Accuracy.}
H\textsuperscript{2}Flow has a lower Scene-min$_{20}$ADE than MoFlow in 69\% of the test plays (by 0.074m on average); its gain is 6.0\% in the fifth of plays that regroup least in the true future and 4.1\% in the fifth that regroups most (2.8\% per agent in both).

\subsection{Sensitivity Analysis}
\label{sec:sensitivity}
\begin{table}[t]
  \centering
  \caption{Sensitivity of H\textsuperscript{2}Flow on NBA, with MoFlow for the settings it shares: min$_{K'}$ADE/Scene-min$_{K'}$ADE (m) of the $K'$ top-scored hypotheses. Shaded: default. Best in bold, second underlined per row.}
  \label{tab:sens}
  \footnotesize
  \setlength{\tabcolsep}{3pt}
  \newcommand{\dflt}{\cellcolor{oursbg}}
  \newcommand{\val}[1]{\emph{#1}}
  \setlength{\aboverulesep}{0.25ex}\setlength{\belowrulesep}{0.35ex}
  \begin{adjustbox}{max width=\columnwidth}
  \begin{tabular}{@{}lcccc@{}}
    \toprule
    Hypotheses $K'$ & \val{1} & \val{5} & \val{10} & \dflt\val{20} \\
    \quad H\textsuperscript{2}Flow & 1.788/1.675 & 1.179/1.451 & \underline{0.908}/\underline{1.395} & \dflt\textbf{0.672}/\textbf{1.373} \\
    \quad MoFlow & 1.885/1.800 & 1.212/1.541 & \underline{0.928}/\underline{1.479} & \textbf{0.692}/\textbf{1.446} \\
    \midrule
    Group sizes $\mathcal{S}$ & \val{$\{3\}$} & \val{$\{2,3\}$} & \dflt\val{$\{2,3,5\}$} & \val{$\{2,3,8\}$} \\
    & 0.674/\textbf{1.367} & \textbf{0.670}/1.379 & \dflt\underline{0.672}/\underline{1.373} & 0.675/1.379 \\
    \midrule
    Euler steps & \dflt\val{10} & \val{20} & & \\
    \quad H\textsuperscript{2}Flow & \dflt\textbf{0.672}/\textbf{1.373} & 0.673/1.374 & & \\
    \quad MoFlow & \textbf{0.692}/\textbf{1.446} & \textbf{0.692}/1.447 & & \\
    \midrule
    Temperature $\kappa$ & \val{0.05} & \dflt\val{0.1} & \val{0.5} & \\
    & 0.676/1.381 & \dflt\textbf{0.672}/\underline{1.373} & \underline{0.674}/\textbf{1.372} & \\
    \midrule
    Decoder blocks $B$ & \val{2} & \dflt\val{4} & & \\
    & 0.677/1.382 & \dflt\textbf{0.672}/\textbf{1.373} & & \\
    \bottomrule
  \end{tabular}
  \end{adjustbox}
\end{table}

\noindent\textbf{Number of Hypotheses.}
Ranked by their scores (per agent for min$_{K'}$ADE, by the mean over agents for Scene-min$_{K'}$ADE), the top five hypotheses keep Scene-min$_{K'}$ADE within 6\% of all 20, while min$_{K'}$ADE rises by 75\% (\cref{tab:sens}); MoFlow, which also scores its hypotheses, is less accurate at every $K'$.
Joint accuracy thus concentrates in the top-ranked hypotheses, and the top scene hypothesis beats each agent's own top one (1.675 \vs 1.788): pooled scores rank futures more reliably.

\noindent\textbf{Other Settings.}
With learned members, group sizes and temperature matter little: both errors stay within 0.006 of the spread of three runs.
Twenty Euler steps bring no gain to either model, whereas two decoder blocks, each rebuilding the groups, raise Scene-min$_{20}$ADE beyond the spread (1.382).

\section{Conclusions}
\label{sec:conclusion}

We introduce hypothesis-conditioned groups for multi-agent trajectory forecasting: every hypothesis forms the groups of the future it predicts.
H\textsuperscript{2}Flow lets the motion and the groups of each hypothesis evolve together during flow matching and improves on MoFlow from single agents to whole scenes in three sports.
Our findings suggest treating interaction structure as part of the predicted future: a forecaster's own estimates of the future can define the groups of each hypothesis, and scoring hypotheses on groups trains them to be right for several agents together.

\noindent\textbf{Limitations.}
H\textsuperscript{2}Flow fixes the sizes of its groups and assumes a fixed set of fully observed agents; groups of dynamic size, partial observations, and agents that enter or leave the scene are left to future work.
It sees only past trajectories, while human motion is inherently stochastic and driven by intentions and context that these trajectories do not reveal.
Its groups are inferred from predicted motion and may miss latent or implicit interactions, such as distant agents coordinating through shared intentions.

{
    \small
    \bibliographystyle{ieeenat_fullname}
    \bibliography{main}
}

\ifwithsupp
\clearpage
\makeatletter\def\theHALG@line{\thealgorithm.\arabic{ALG@line}}\makeatother  %
\twocolumn[{\centering\Large\textbf{Appendix}\par}\vspace{1.0em}]
\appendix

\noindent\textbf{Overview.}
\suppapp{This document completes the main paper}{This appendix completes the main text} with the details of the model and the experiments and with the results left out for space.
Each part below names the sections of \suppapp{the main paper}{the main text} that it extends.
\begin{itemize}
  \setlength{\itemsep}{1pt}
  \item \textbf{\ref{sec:supp_method}. Model details} (\suppapp{Sec.~3}{\cref{sec:method}}): the notation (\cref{tab:supp_notation}), the layers of the encoder and the decoder (\cref{fig:supp_dec_flow}), the flow parameterization and the losses, and the training and sampling algorithms (\cref{alg:train,alg:sample}, \cref{fig:supp_flow_loop}).
  \item \textbf{\ref{sec:supp_impl}. Experimental setup} (\suppapp{Secs.~4.1--4.3}{Secs.~\ref{sec:data}--\ref{sec:baselines}}): the datasets (\cref{tab:supp_data}, \cref{fig:supp_data}), the joint metrics G3, Team, and Scene, the training settings (\cref{tab:supp_hparams}), and the baselines.
  \item \textbf{\ref{sec:supp_quant}. Additional quantitative results} (\suppapp{Secs.~4.4, 4.6, and 4.8}{Secs.~\ref{sec:quant}, \ref{sec:ablation}, and \ref{sec:sensitivity}}): the error over the horizon (\cref{fig:supp_horizon,fig:supp_gains}, \cref{tab:supp_horizon}), the significance of the gains (\cref{tab:supp_signif}), joint accuracy on ETH-UCY (\cref{tab:supp_ped_joint}), the ablation on NFL and soccer and, on NBA, without the endpoint and $\mathcal{L}_{\mathrm{grp}}$ (\cref{tab:supp_ablation,tab:supp_ablation_chain}, \cref{fig:supp_ablation_pipeline}), and the number and diversity of the hypotheses (\cref{fig:supp_mink}, \cref{tab:supp_diversity}).
  \item \textbf{\ref{sec:complexity}. Computational cost}: the training and sampling time of all models on the three sports (\cref{tab:supp_cost}), their accuracy against cost on NBA (\cref{fig:supp_cost}), and the growth of the sampling cost with the numbers of agents and hypotheses (\cref{fig:supp_scaling}).
  \item \textbf{\ref{sec:supp_groups}. Flow steps and hypothesis groups} (\suppapp{Sec.~4.7}{\cref{sec:analysis}}): the states, endpoints, and groups of the hypotheses over the flow steps (\cref{fig:supp_flow_process,fig:supp_flow_groups}), and statistics of the groups (\cref{tab:supp_groupstats}).
  \item \textbf{\ref{sec:supp_qual}. Additional qualitative results} (\suppapp{Sec.~4.5}{\cref{sec:qual}}): \suppapp{the selection rules, comparisons with MoFlow including failure cases (\cref{fig:supp_samples,fig:supp_transitions,fig:supp_failures}), and, on the three sports and on ETH-UCY, the best joint hypothesis of every model (\cref{fig:supp_pred_gt,fig:supp_pred_gt_nfl,fig:supp_pred_gt_soccer,fig:supp_pred_gt_ethucy}), the hypothesis groups (\cref{fig:supp_hyp_groups,fig:supp_hyp_groups_scales,fig:supp_hyp_groups_sports,fig:supp_hyp_groups_ethucy}), and}{the selection rules, the best joint hypothesis of every model on NFL, soccer, and ETH-UCY (\cref{fig:supp_pred_gt_nfl,fig:supp_pred_gt_soccer,fig:supp_pred_gt_ethucy}) with failure cases against MoFlow (\cref{fig:supp_failures}), the hypothesis groups (\cref{fig:supp_hyp_groups_scales,fig:supp_hyp_groups_sports,fig:supp_hyp_groups_ethucy}), and} all 20 futures of two agents (\cref{fig:qual_k_futures,fig:supp_k_futures,fig:supp_k_futures_nfl,fig:supp_k_futures_soccer,fig:supp_k_futures_ethucy}).
\end{itemize}
\suppapp{Tables, figures, equations, and algorithms of this document are numbered S1, S2, \ldots; items of the main paper are cited with their numbers there.}{}

\section{Model Details}
\label{sec:supp_method}
This section gives the equations of every module of \mainpaper{fig:overview}{Fig.~2} as implemented in our code, and \cref{tab:supp_notation} lists the symbols \suppapp{of both documents}{of the main text and the appendix}.
Linear maps $W$ include a bias, $\mathrm{FFN}(x)=W_2\,\mathrm{ReLU}(W_1x)$ is a two-layer network, $\mathrm{PE}$ is a sinusoidal encoding, $\sigma$ is the sigmoid, and $\mathrm{sg}[\cdot]$ stops the gradient.
Every attention or feed-forward sublayer $f$ is followed by a residual connection and layer normalization (post-norm), $x\leftarrow\mathrm{LN}(x+f(x))$, and $\mathrm{TF}$ is a transformer layer: a multi-head self-attention sublayer and a feed-forward sublayer with $4D$ hidden channels.
Attention layers have 8 heads of $d_h=D/8$ channels, with $D=128$; the equations show one head, whose outputs are concatenated over the heads and projected linearly.

\begin{table}[t]
  \centering
  \caption{Notation.}
  \label{tab:supp_notation}
  \footnotesize
  \setlength{\tabcolsep}{3pt}
  \begin{tabular}{@{}lp{0.74\columnwidth}@{}}
    \toprule
    Symbol & Meaning \\
    \midrule
    $N$; $i,j$ & number of agents; agent indices \\
    $K$; $k$ & number of hypotheses; hypothesis index \\
    $K'$ & number of hypotheses kept (\cref{sec:supp_mink}) \\
    $T_p$, $T_f$; $\tau$ & observed and future frames; frame index \\
    $E$; $e$ & number of hyperedges per segment; hyperedge \\
    $\mathcal{S}$; $s$ & set of group sizes; group size \\
    $M$; $m$ & number of horizon segments; segment index \\
    $B$; $b$ & number of decoder blocks; block index \\
    $L$; $D$ & number of layers of each encoder transformer; hidden channels \\
    $t$; $\Delta t$ & flow time; Euler step size \\
    $C$; $c_{i,\tau}$ & context, $N\times T_p\times 6$; its features of agent $i$ at frame $\tau$ \\
    $Y$, $Y^0$, $Y^t$ & true future, source sample, and path point, $N\times 2T_f$ per hypothesis \\
    $S$, $\hat Y$, $\zeta$ & predicted endpoints, self-conditioning estimate, logits \\
    $D_\theta$; $d_{k,i}$; $\mathcal{G}$ & decoder; error of hypothesis $k$ for agent $i$; groups of a joint metric \\
    $x_i$, $\dot x_i$ & position and velocity of agent $i$ at the last observed frame \\
    $u_{i,\tau}$, $\rho_i$ & frame embedding; role embedding \\
    $n^0_i$, $n^{\mathrm{pair}}_i$, $n^{(s)}_i$ & agent features after the embedding, the pairwise, and the group transformer of size $s$ \\
    $r_{ij}$, $h_e$, $\bar h_i$ & pair relation, hyperedge feature, mean feature of the hyperedges of agent $i$ \\
    $\ell_{ij}$; $d_h$ & attention logit; channels per head \\
    $F$, $\Gamma$ & encoder agent features and group tokens \\
    $H^{(s)}$, $H^{b,m}_k$ & encoder incidence of size $s$; groups of hypothesis $k$ in block $b$ and segment $m$ \\
    $\alpha$, $\delta^m_{k,ij}$ & affinity; mean distance of agents $i,j$ in the estimated future \\
    $\bar z_i$, $\bar g_{k,i}$ & centered unit projections of the encoder and decoder builders \\
    $\alpha^{(s)}_j$, $\bar\alpha^{(s)}_j$ & $s$-th largest affinity of agent $j$; cut between the members and the rest \\
    $\tilde H$, $\sigma^{(s)}_{ij}$ & straight-through incidence; soft incidence \\
    $G$, $\gamma^{b,m}_{k,e}$ & decoder tokens; hypothesis group tokens \\
    $U_{k,i}$, $c_{\mathrm{in}}(t)$ & embedding of the path point (and estimate); input scale of the path point \\
    $\omega_k$, $\nu_i$, $\phi_t$, $\psi_m$ & embeddings of hypothesis, agent, flow time, and segment \\
    $\kappa$; $w$, $\beta$ & straight-through temperature; weight of the future term, its parameter \\
    $\mu_b$, $\eta_b$ & scale and shift of the time modulation in block $b$ \\
    $\tilde Y$; $\sigma_d$ & normalized future; standard deviation of the data \\
    $y_{\min}$, $y_{\max}$ & bounds of the training futures \\
    $p_{\mathrm{drop}}(t)$; $k^*_i$ & drop probability of the path-point embedding; winner of agent $i$ \\
    $e_k$, $\epsilon_{k,e}$ & group of hypothesis $k$ at hyperedge index $e$; its mean error \\
    $\lambda_q$, $J$ & weight of $\mathcal{L}_{\mathrm{grp}}$ at epoch $q$; number of epochs \\
    \bottomrule
  \end{tabular}
\end{table}

\subsection{Scene Hypergraph Encoder}
\label{sec:supp_arch}

\noindent\textbf{Agent Embedding.}
Frame $\tau$ of agent $i$ holds the 6 features $c_{i,\tau}$ of the context $C$ (\mainpaper{sec:background}{Sec.~3.1}).
They are embedded with the frame index and flattened over the $T_p$ frames,
\begin{equation}
\begin{aligned}
  u_{i,\tau}&=W_t\big[W_c\,c_{i,\tau};\,\mathrm{PE}(\tau)\big],\\
  n^0_i&=W_n\big[u_{i,1};\dots;u_{i,T_p}\big]+\rho_i,
\end{aligned}
\label{eq:supp_embed}
\end{equation}
where $\rho_i$ is the learned embedding of the role of agent $i$ (team one, team two, or ball), used on NBA only.
The velocity in the relations of \mainpaper{eq:geo}{Eq.~(5)} is the last observed displacement per frame, $\dot x_i=x_{i,T_p}-x_{i,T_p-1}$.

\noindent\textbf{Pairwise Interaction Transformer.}
Starting from $n^0$ and the relations $r_{ij}$ of \mainpaper{eq:geo}{Eq.~(5)}, each of the $L$ layers lets agent $i$ attend to every agent $j$ through their relation~\cite{lee2024mart}:
\begin{equation}
\begin{aligned}
  \ell_{ij}&=\frac{\big(W_qn_i+W^r_qr_{ij}\big)^{\top}\big(W_kn_j+W^r_kr_{ij}\big)}{\sqrt{d_h}},\\
  n_i&\leftarrow\mathrm{LN}\Big(n_i+\sum_{j=1}^{N}\mathrm{softmax}_j(\ell_{ij})\,v_{ij}\Big),
\end{aligned}
\label{eq:supp_pair_att}
\end{equation}
with the value $v_{ij}=W_vn_j+W^r_vr_{ij}$, followed by $n_i\leftarrow\mathrm{LN}(n_i+\mathrm{FFN}(n_i))$.
The relations are then updated from both directions of the pair and both agents,
\begin{equation}
\begin{aligned}
  r_{ij}&\leftarrow\mathrm{LN}\big(r_{ij}+\mathrm{FFN}([r_{ij};\,r_{ji};\,n_j;\,n_i])\big),\\
  r_{ij}&\leftarrow\mathrm{LN}\big(r_{ij}+\mathrm{FFN}(r_{ij})\big).
\end{aligned}
\label{eq:supp_pair_upd}
\end{equation}
After $L$ layers, the agent features are $n^{\mathrm{pair}}_i$.
All feed-forward networks of the encoder have $2D$ hidden channels, and the encoder uses no dropout on the team sports.

\noindent\textbf{Hyperedge Builder.}
\label{sec:supp_builder}
The builder projects the agent embeddings, removes their mean over the agents of the scene, and normalizes them to unit length:
\begin{equation}
\begin{aligned}
  z_i&=W_zn^0_i,\qquad\bar z_i=\frac{z_i-\langle z\rangle}{\lVert z_i-\langle z\rangle\rVert_2},\\
  \alpha_{ij}&=\bar z_i^{\top}\bar z_j,
\end{aligned}
\label{eq:supp_aff}
\end{equation}
where $\langle z\rangle=\frac{1}{N}\sum_jz_j$ and $W_z$ is initialized to the identity.
The centering removes the component that all embeddings share, which would otherwise give every pair a high affinity.
As $\alpha_{jj}=1$ is the largest affinity of agent $j$, every agent belongs to the hyperedges it anchors.
With the $s$-th largest affinity $\alpha^{(s)}_j$ of agent $j$ and the midpoint $\bar\alpha^{(s)}_j=\frac{1}{2}(\alpha^{(s)}_j+\alpha^{(s+1)}_j)$ between the last member and the first non-member, the incidence of size $s$ is
\begin{equation}
\begin{aligned}
  \tilde H^{(s)}_{i,j}&=\mathds{1}\big[\alpha_{ij}\ge\alpha^{(s)}_j\big]+\sigma^{(s)}_{ij}-\mathrm{sg}\big[\sigma^{(s)}_{ij}\big],\\
  \sigma^{(s)}_{ij}&=\sigma\big((\alpha_{ij}-\bar\alpha^{(s)}_j)/\kappa\big),
\end{aligned}
\label{eq:supp_ste}
\end{equation}
with $\kappa=0.1$.
Its value is the hard incidence $H^{(s)}$ of \suppapp{the main paper}{\cref{sec:encoder}}, and its gradient is that of the soft incidence: at a fixed cut, $\partial\tilde H^{(s)}_{i,j}/\partial\alpha_{ij}=\sigma^{(s)}_{ij}(1-\sigma^{(s)}_{ij})/\kappa$, which is largest for the agents closest to the cut.
If $s$ exceeds the number of agents, the hyperedge holds all of them.
Every agent anchors one hyperedge of every size, so $E=N|\mathcal{S}|$ (33 on NBA, 69 on NFL and soccer); as all sizes rank the same affinities, the hyperedges of one anchor are nested.

\noindent\textbf{Group Interaction Transformers.}
The transformer of size $s$ embeds every hyperedge $e$ from its members,
\begin{equation}
  h_e=\mathrm{FFN}_h\Big(\frac{1}{|e|}\sum_{i=1}^{N}\tilde H^{(s)}_{i,e}\,n^0_i\Big).
  \label{eq:supp_hinit}
\end{equation}
In each of its $L$ layers, every agent adds the mean feature $\bar h_i=\sum_e\tilde H^{(s)}_{i,e}h_e/\sum_e\tilde H^{(s)}_{i,e}$ of its hyperedges to its query, key, and value~\cite{lee2024mart}:
\begin{equation}
\begin{aligned}
  \ell_{ij}&=\frac{\big(W_qn_i+W^h_q\bar h_i\big)^{\top}\big(W_kn_j+W^h_k\bar h_j\big)}{\sqrt{d_h}},\\
  n_i&\leftarrow\mathrm{LN}\Big(n_i+\sum_{j=1}^{N}\mathrm{softmax}_j(\ell_{ij})\,v_j\Big),
\end{aligned}
\label{eq:supp_hatt}
\end{equation}
with the value $v_j=W_vn_j+W^h_v\bar h_j$, followed by the feed-forward sublayer.
Every hyperedge is then updated from the mean of its members,
\begin{equation}
  h_e\leftarrow\mathrm{LN}\Big(h_e+\mathrm{FFN}\Big(\Big[h_e;\,\frac{1}{|e|}\sum_{i=1}^{N}\tilde H^{(s)}_{i,e}\,n_i\Big]\Big)\Big),
  \label{eq:supp_hupd}
\end{equation}
followed by $h_e\leftarrow\mathrm{LN}(h_e+\mathrm{FFN}(h_e))$.
Through the straight-through incidence in \cref{eq:supp_hinit,eq:supp_hatt,eq:supp_hupd}, the forecasting loss trains the builder.
Each size runs its own $L$ layers from $n^0$, and the final agent features $n^{(s)}_i$ and hyperedge features $h_e$ of all sizes are fused into $F$ and $\Gamma$ (\mainpaper{sec:encoder}{Sec.~3.3.1}).

\subsection{Hypothesis-Conditioned Decoder}
\label{sec:supp_decoder}

\begin{figure*}[t]
  \centering
  \includegraphics[width=\textwidth]{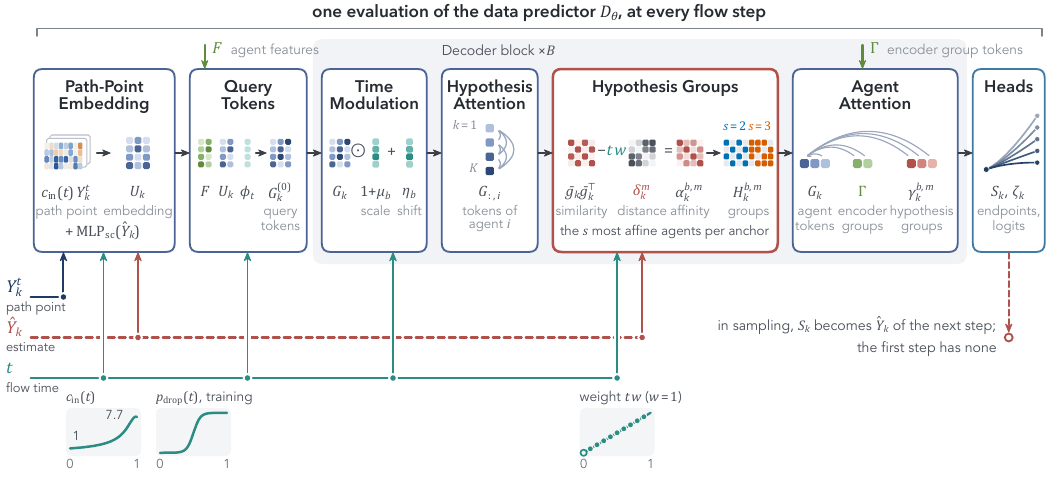}
  \caption{Where the inputs of one decoder evaluation enter $D_\theta$ (\cref{sec:supp_decoder}). Top: its stages and their symbols. Lanes: the path point $Y^t$, the flow time $t$, and the estimate $\hat Y_k$ (dashed; absent at the first step), each linked to where it enters. Charts: the schedules $c_{\mathrm{in}}(t)$, $p_{\mathrm{drop}}(t)$, and $t\,w$. Glyphs are schematic.}
  \label{fig:supp_dec_flow}
\end{figure*}

\cref{fig:supp_dec_flow} shows where the three inputs of one decoder evaluation enter: the path point $Y^t$, the estimate $\hat Y$, and the flow time $t$.
The path point and the estimate enter the tokens, and the estimate also sets the future term of the groups.
The flow time scales the path point, conditions the tokens of every block, weights the future term, and turns the endpoint into a velocity.

\noindent\textbf{Flow Time and Path Point.}
The flow time is embedded as $\phi_t=\mathrm{FFN}_t(\mathrm{PE}(1000\,t))$.
The embedding of the path point mixes the hypotheses and the agents with two transformer layers of 4 heads,
\begin{equation}
\begin{aligned}
  U_{k,i}&=\mathrm{MLP}_U\big(c_{\mathrm{in}}(t)\,Y^t_{k,i}\big)+\omega_k+\nu_i,\\
  U_{:,i}&\leftarrow\mathrm{TF}^U_K(U_{:,i}),\qquad U_{k,:}\leftarrow\mathrm{TF}^U_N(U_{k,:}),
\end{aligned}
\label{eq:supp_path}
\end{equation}
where $\mathrm{MLP}_U$ maps $2T_f\to D\to D\to D$ with ReLU, $c_{\mathrm{in}}$ is the input scale of \cref{eq:supp_cin}, $\mathrm{TF}^U_K$ attends over the $K$ hypotheses of agent $i$, and $\mathrm{TF}^U_N$ over the $N$ agents of hypothesis $k$.
In training, the embedding of a scene is set to zero with probability $p_{\mathrm{drop}}(t)=\sigma(20(t-\frac{1}{2}))$, which drops half of the scenes at $t=0.5$ and over 99\% above $t=0.75$.
Late in the path, the decoder thus learns to predict the endpoint from the context and the estimate.
Self-conditioning then adds the embedded estimate,
\begin{equation}
  U_{k,i}\leftarrow U_{k,i}+\mathrm{MLP}_{\mathrm{sc}}(\hat Y_{k,i}),
  \label{eq:supp_sc}
\end{equation}
where $\mathrm{MLP}_{\mathrm{sc}}$ maps $2T_f\to D\to D$ with ReLU; without an estimate, nothing is added.
The query tokens $G^{(0)}_{k,i}$ follow \mainpaper{sec:hypgroups}{Sec.~3.3.2}, where $\mathrm{MLP}_1$ ($3D\to D\to D$) and $\mathrm{MLP}_2$ ($D\to D\to D$) have a layer normalization and a ReLU between their two layers, and $\omega_k$ and $\nu_i$ are the embeddings of \cref{eq:supp_path}.

\noindent\textbf{Decoder Block.}
Block $b$ updates the tokens $G=G^{(b-1)}$ in four steps.
First, the flow time modulates every token,
\begin{equation}
\begin{aligned}
  G_{k,i}&\leftarrow(1+\mu_b)\odot G_{k,i}+\eta_b,\\
  [\mu_b;\,\eta_b]&=W^b_t\,\mathrm{SiLU}(\phi_t),
\end{aligned}
\label{eq:supp_mod}
\end{equation}
with $W^b_t$ initialized to zero, so every block starts from the identity.
Second, the $K$ hypotheses of every agent attend to each other, $G_{:,i}\leftarrow\mathrm{TF}^b_K(G_{:,i})$.
Third, every hypothesis forms its groups with the builder of block $b$ applied to its tokens, with the affinity of \mainpaper{eq:futaff}{Eq.~(7)}:
\begin{equation}
\begin{aligned}
  g_{k,i}&=W^b_zG_{k,i},\qquad\bar g_{k,i}=\frac{g_{k,i}-\langle g_k\rangle}{\lVert g_{k,i}-\langle g_k\rangle\rVert_2},\\
  \alpha^{b,m}_{k,ij}&=\bar g_{k,i}^{\top}\bar g_{k,j}-t\,w\,\delta^m_{k,ij},
\end{aligned}
\label{eq:supp_decaff}
\end{equation}
where $\langle g_k\rangle$ averages over the agents of hypothesis $k$, and $w=\frac{1}{10}\mathrm{softplus}(\beta)$, with a learned $\beta$ initialized so that $w=0.1$.
The distance $\delta^m_{k,ij}$ is measured between absolute positions in meters; as $\delta^m_{k,jj}=0$, every agent stays in the hyperedges it anchors.
\Cref{eq:supp_ste} with this affinity gives the incidence $\tilde H^{b,m}_k$, and the group tokens pool their members,
\begin{equation}
  \gamma^{b,m}_{k,e}=\mathrm{LN}\Big(W^b_h\frac{\sum_i\tilde H^{b,m}_{k,i,e}\,G_{k,i}}{\sum_i\tilde H^{b,m}_{k,i,e}}\Big)+\psi_m,
  \label{eq:supp_gtok}
\end{equation}
with the segment embedding $\psi_m$ initialized to zero.
Without an estimate (at the first flow step and in the training batches without self-conditioning), the future term is absent, and one set of $E$ groups without segment embedding replaces the $M$ segment sets.
Fourth, the agents of hypothesis $k$ attend to the $N+(M+1)E$ tokens $\mathcal{X}_k=[G_{k,:};\,\Gamma;\,\gamma^{b,1}_k;\,\gamma^{b,2}_k]$ in the transformer layer $\mathrm{TF}^b_N$ ($\mathrm{AgentAttn}_b$ in \mainpaper{eq:block}{Eq.~(10)}).
The attention of agent $i$ normalizes over the agent and group tokens together,
\begin{equation}
  o_{k,i}=\sum_{x\in\mathcal{X}_k}\mathrm{softmax}_x\Big(\frac{(W_qG_{k,i})^{\top}W_kx}{\sqrt{d_h}}\Big)\,W_vx,
  \label{eq:supp_agentattn}
\end{equation}
and only the $N$ agent outputs are kept.
All decoder layers use dropout 0.1.

\noindent\textbf{Output Heads.}
The endpoint and the logits read the tokens $G^{(B)}$ of the last block (\mainpaper{eq:gf}{Eq.~(11)}).
$\mathrm{MLP}_{\mathrm{reg}}$ maps $D\to128\to256\to2T_f$, $\mathrm{MLP}_{\mathrm{cls}}$ maps $D\to128\to128\to1$, and $\mathrm{MLP}_{\mathrm{grp}}$ maps $D\to D\to2T_f$, all with ReLU.
The output layer of $\mathrm{MLP}_{\mathrm{grp}}$ is initialized to zero, so training starts from the per-agent endpoint.
In the group-factorized endpoint (\mainpaper{eq:gf}{Eq.~(11)}), the denominator counts the groups that contain agent $i$, at least the $M|\mathcal{S}|$ it anchors, and the straight-through incidence carries the gradient of the endpoint error to the builder of the last block.

\subsection{Flow Parameterization and Losses}
\label{sec:supp_flow}

\noindent\textbf{Coordinates.}
The network works in normalized coordinates, in which every coordinate of the future offsets lies in $[-1,1]$:
\begin{equation}
  \tilde Y=2\,\frac{Y-y_{\min}}{y_{\max}-y_{\min}}-1,
  \label{eq:supp_norm}
\end{equation}
with the smallest and largest coordinate $y_{\min}$ and $y_{\max}$ of the training futures.
The noise, the path, the endpoints $S$, and the estimate $\hat Y$ live in these coordinates, whereas the errors $d_{k,i}$ of all losses and the distances $\delta^m_{k,ij}$ are computed in meters after inverting \cref{eq:supp_norm}.

\noindent\textbf{Input Scale and Flow Time.}
The path point enters the decoder scaled to unit variance,
\begin{equation}
  c_{\mathrm{in}}(t)=\big(\sigma_d^2t^2+(1-t)^2\big)^{-1/2},
  \label{eq:supp_cin}
\end{equation}
the inverse standard deviation of $Y^t=(1-t)Y^0+t\tilde Y$ for unit-variance noise and data of standard deviation $\sigma_d$, with $\sigma_d=0.13$ (0.0663 on SDD).
$D_\theta$ outputs the endpoint directly, and the losses carry no weight over $t$.
Training draws $t=\sigma(u)$ with $u\sim\mathcal{N}(-0.5,1.5^2)$, the logit-normal density
\begin{equation}
  p(t)=\frac{1}{1.5\sqrt{2\pi}\,t(1-t)}\exp\Big(-\frac{(\mathrm{logit}\,t+0.5)^2}{2\cdot1.5^2}\Big),
  \label{eq:supp_tdens}
\end{equation}
whose median $\sigma(-0.5)=0.38$ favors the noisier half of the path.
Sampling takes 10 Euler steps of equal size, at $t_n=n/10$ for $n=0,\dots,9$.

\noindent\textbf{Losses.}
With the winner $k^*_i=\argmin_kd_{k,i}$ of agent $i$, the per-agent losses of \mainpaper{sec:bg_moflow}{Sec.~3.2} are
\begin{equation}
\begin{aligned}
  \mathcal{L}_{\mathrm{agent}}&=\frac{1}{N}\sum_{i=1}^{N}d_{k^*_i,i},\\
  \mathcal{L}_{\mathrm{cls}}&=-\frac{1}{N}\sum_{i=1}^{N}\log\frac{\exp\zeta_{k^*_i,i}}{\sum_{k=1}^{K}\exp\zeta_{k,i}}.
\end{aligned}
\label{eq:supp_agent}
\end{equation}
$\mathcal{L}_{\mathrm{grp}}$ (\mainpaper{eq:grp}{Eq.~(13)}) uses the hard incidence of the last block with the gradient stopped, $H_{k,:,e}=\mathrm{sg}[H^{B,m}_{k,:,e}]$, over its $ME$ hyperedge indices ($E$ in the batches without an estimate).
It scores the groups of both segments by the error over the whole horizon.
The implementation sums the displacements over the $T_f$ frames in the regression terms, which scales $d_{k,i}$ by $T_f$ there, so the loss of a batch $\mathcal{B}$ of scenes is
\begin{equation}
  \mathcal{L}=\frac{1}{|\mathcal{B}|}\sum_{\mathcal{B}}\Big[T_f\big(\mathcal{L}_{\mathrm{agent}}+\lambda_q\,\mathcal{L}_{\mathrm{grp}}\big)+\mathcal{L}_{\mathrm{cls}}\Big].
  \label{eq:supp_loss}
\end{equation}
This scaling leaves the winners unchanged and weights the regression terms by $T_f$ (20 on the team sports, 12 on ETH-UCY and SDD) relative to $\mathcal{L}_{\mathrm{cls}}$.
For self-conditioning, a batch first estimates $\hat Y=\mathrm{sg}[S]$ with probability $\frac{1}{2}$, from the same $t$ and $Y^t$ and without an estimate; the other batches use none.
We optimize with AdamW, with a learning rate that rises linearly over the first 5\% of the iterations and then decays to $10^{-6}$ along a cosine, and clip the gradients to norm 1.

\subsection{Training and Sampling Algorithms}
\label{sec:supp_algo}
\cref{alg:train,alg:sample} give one training step and the sampling of $K$ hypotheses (\suppapp{Secs.~3.4 and 3.5 of the main paper}{\cref{sec:train_sample,sec:inference}}), and \cref{fig:supp_flow_loop} illustrates both.
In training, $Y^t$ is a point of the linear path, and the estimate comes from a first pass at the same $t$ and $Y^t$, so the decoder learns to use an estimate without unrolling the sampler.
In sampling, $Y^t$ is the state of the Euler solver and the estimate is the endpoint of the previous step; each step moves the state $\Delta t/(1-t)$ of the way toward the new endpoint, so the last step lands on it.

\begin{figure*}[t]
  \centering
  \includegraphics[width=\textwidth]{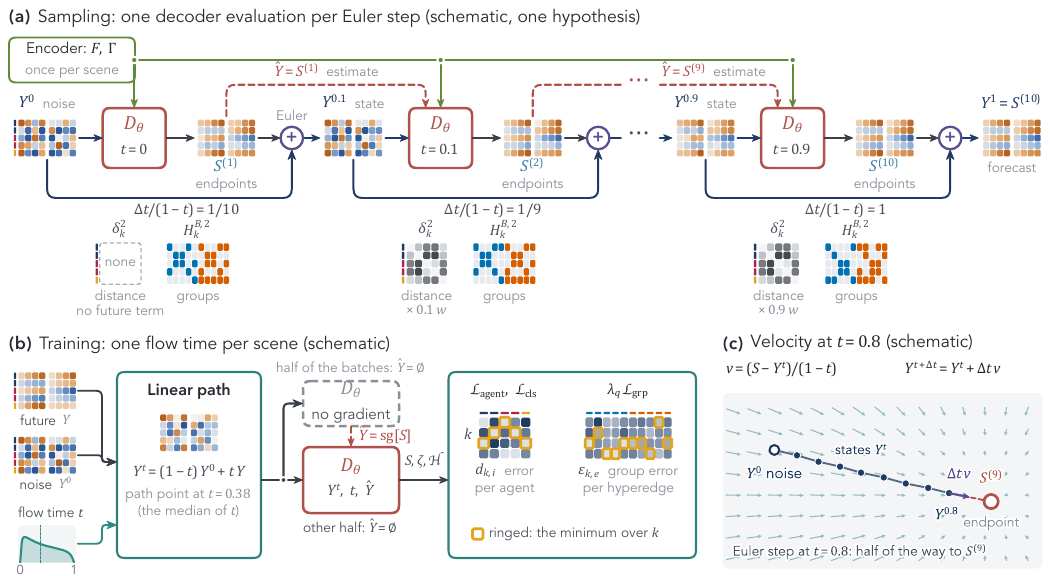}
  \caption{The hypothesis-conditioned decoder in sampling and training. Schematic, on a toy scene: five agents, four frames, $K=4$; a future is drawn as its $x$ and $y$ displacements. (a) Sampling (\cref{alg:sample}), steps 1, 2, and 10: the state moves $\Delta t/(1-t)$ of the way toward the predicted endpoints $S^{(n)}$, which become the next estimate $\hat Y$; below, the distances $\delta^2_k$ and groups $H^{B,2}_k$ formed from it. (b) Training (\cref{alg:train}): the flow-time density (dotted: median); in half of the batches, a first pass without gradients gives the estimate; matrices: errors $d_{k,i}$ per agent and $\epsilon_{k,e}$ per hyperedge (ring: regressed hypothesis). (c) One coordinate pair of (a) (agent 3, frame 2): its states, the field $v$ of its endpoint $S^{(9)}$ at $t=0.8$, and the Euler step, which covers half of the remaining way.}
  \label{fig:supp_flow_loop}
\end{figure*}

\begin{algorithm}[t]
  \caption{One training step of H\textsuperscript{2}Flow.}
  \label{alg:train}
  \small
  \begin{algorithmic}[1]
    \Require context $C$, normalized future $\tilde Y$ (\cref{eq:supp_norm}), epoch $q$
    \State $(F,\Gamma)\gets\mathrm{Enc}_\theta(C)$
    \State $t\sim\mathrm{LogitNormal}(-0.5,1.5)$,\; $Y^0\sim\mathcal{N}(\mathbf{0},\mathbf{I})$
    \State $Y^t_k\gets(1-t)\,Y^0+t\,\tilde Y$ for $k=1,\dots,K$ \Comment{noise shared by the hypotheses}
    \State $\hat Y\gets\varnothing$
    \If{$u<\frac{1}{2}$ with $u\sim\mathcal{U}(0,1)$} \Comment{self-conditioning}
      \State $\hat Y\gets\mathrm{sg}\big[S\big]$ with $(S,\cdot,\cdot)=\mathrm{Dec}_\theta(Y^t,t,\varnothing;F,\Gamma)$
    \EndIf
    \State $(S,\zeta,\mathcal{H})\gets\mathrm{Dec}_\theta(Y^t,t,\hat Y;F,\Gamma)$ \Comment{groups $H^{b,m}_k$ in every block}
    \State $d_{k,i}\gets$ error of $S_{k,i}$,\; $k^*_i\gets\argmin_k d_{k,i}$
    \State $\mathcal{L}\gets T_f\big(\mathcal{L}_{\mathrm{agent}}+\lambda_q\,\mathcal{L}_{\mathrm{grp}}\big)+\mathcal{L}_{\mathrm{cls}}$ \Comment{\cref{eq:supp_loss}}
    \State update $\theta$ with $\nabla_\theta\mathcal{L}$
  \end{algorithmic}
\end{algorithm}

\begin{algorithm}[t]
  \caption{Sampling $K$ hypotheses with $n$ Euler steps.}
  \label{alg:sample}
  \small
  \begin{algorithmic}[1]
    \Require context $C$, number of steps $n$
    \State $(F,\Gamma)\gets\mathrm{Enc}_\theta(C)$ \Comment{once per scene}
    \State $Y^0\sim\mathcal{N}(\mathbf{0},\mathbf{I})$,\; $Y^0_k\gets Y^0$ for all $k$,\; $\hat Y\gets\varnothing$,\; $\Delta t\gets 1/n$
    \For{$t=0,\Delta t,\dots,1-\Delta t$}
      \State $(S,\zeta,\mathcal{H})\gets\mathrm{Dec}_\theta(Y^t,t,\hat Y;F,\Gamma)$ \Comment{groups from tokens and $\hat Y$}
      \State $Y^{t+\Delta t}_k\gets Y^t_k+\Delta t\,(S_k-Y^t_k)/(1-t)$ for all $k$
      \State $\hat Y\gets S$
    \EndFor
    \State map $S$ back to meters (\cref{eq:supp_norm})
    \State \Return trajectories $x_i+S_{k,i}$ and scores $\mathrm{softmax}_k(\zeta_{:,i})$
  \end{algorithmic}
\end{algorithm}

\section{Experimental Setup}
\label{sec:supp_impl}
This section completes the setup of \suppapp{Secs.~4.1--4.3 of the main paper}{Secs.~\ref{sec:data}--\ref{sec:baselines}}.

\subsection{Datasets}
\label{sec:supp_data}

\begin{table}[t]
  \centering
  \caption{Datasets: agents per scene (ETH-UCY: mean and maximum per test sequence; SDD: ego + neighbors), training and test scenes (ETH-UCY: trajectories; SDD: samples), and mean displacement of an agent over the future (4.0s; ETH-UCY and SDD: 4.8s).}
  \label{tab:supp_data}
  \footnotesize
  \setlength{\tabcolsep}{3pt}
  \begin{adjustbox}{max width=\columnwidth}
  \begin{tabular}{lcccc}
    \toprule
    Dataset & Agents & Train & Test & Motion \\
    \midrule
    NBA & 11 (10 + ball) & 32,500 & 12,500 & 5.8m \\
    NFL & 23 (22 + ball) & 23,127 & 3,548 & 11.2m \\
    Soccer & 23 (22 + ball) & 32,225 & 684 & 6.7m \\
    \midrule
    ETH & 2.6 (max 5) & 29,809 & 181 & 4.8m \\
    Hotel & 3.5 (max 8) & 29,152 & 1,053 & 2.1m \\
    Univ & 25.7 (max 57) & 9,231 & 24,334 & 2.5m \\
    Zara1 & 3.7 (max 14) & 28,010 & 2,253 & 4.6m \\
    Zara2 & 6.3 (max 14) & 25,507 & 5,833 & 2.5m \\
    \midrule
    SDD & 1 + 1.0 (max 1 + 10) & 8,985 & 2,829 & 127px \\
    \bottomrule
  \end{tabular}
  \end{adjustbox}
\end{table}

\begin{figure*}[t]
  \centering
  \includegraphics[width=\textwidth]{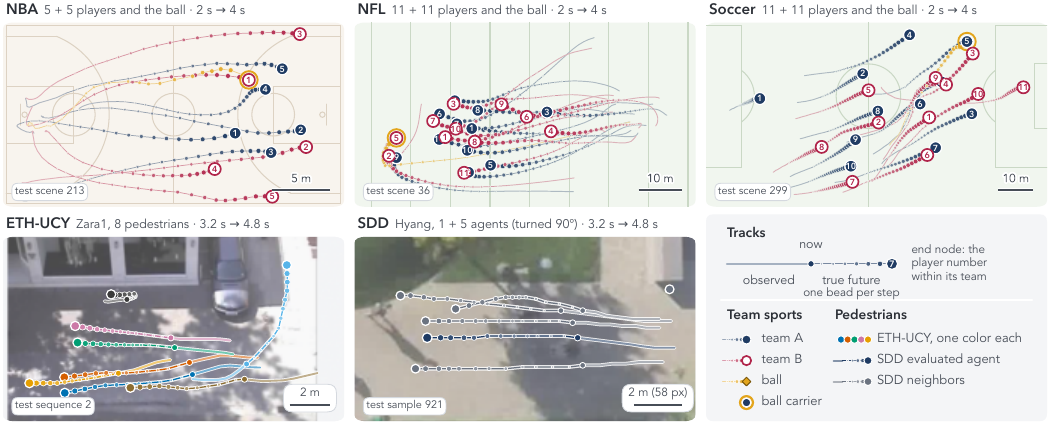}
  \caption{One test scene of each dataset; panel titles give the agents and the horizon. ETH-UCY and SDD lie on their camera frames (SDD in pixels, turned by $90^\circ$). Scenes: on the team sports, the one whose players move farthest among the first 500 test scenes; on Zara1 and SDD, the first with eight pedestrians and with five neighbors.}
  \label{fig:supp_data}
\end{figure*}

\cref{tab:supp_data} summarizes the five datasets, and \cref{fig:supp_data} shows one test scene of each.
Over the future, NFL players move about twice as far as NBA players, and soccer players in between.

\noindent\textbf{NBA} SportVU tracking uses the split of MoFlow~\cite{fu2025moflow}.

\noindent\textbf{NFL} uses the Next Gen Stats tracking of the 2017 season released with Sports-Traj~\cite{xu2025sportstraj}.
Every play with 11 players per team and the ball in all its frames is resampled from 10 to 5Hz and converted from yards to meters.
It is cut into windows of 2s observed and 4s future that start every 1s for training and every 2s for testing, with training and test plays from different games (73 and 18).

\noindent\textbf{Soccer} uses the top-view recordings of one match of SoccerTrack~\cite{scott2022soccertrack} (60 clips of 30s at 30fps).
The center of each bounding box in the 4K drone frame is converted to meters with the pitch width (68m over 2,160 pixels) and resampled to 5Hz.
Windows of 6s start every 0.2s for training and every 0.4s for testing, without crossing a clip (48 and 12 clips).
On NFL and soccer, windows with a tracking glitch (a player step faster than 12m/s or a ball step faster than 40m/s) are dropped, and the agents are ordered by team, with the ball last.

\noindent\textbf{ETH-UCY}~\cite{pellegrini2009eth,lerner2007ucy} uses the leave-one-out split of SocialGAN~\cite{gupta2018socialgan}.
A sequence holds all pedestrians present in the same 20 frames at 2.5Hz, so the test sequences of Univ hold many more pedestrians than those of the other subsets.

\noindent\textbf{SDD}~\cite{robicquet2016sdd} uses the split of TUTR~\cite{shi2023tutr} in pixels, with one evaluated ego per sample and its neighbors as context.

\noindent\textbf{Rendering.}
The figures of \suppapp{this document}{this appendix} draw ETH-UCY on the camera frame of each scene from OpenTraj~\cite{amirian2020opentraj}, mapped onto the ground plane with the homography of the dataset.
On Zara2 and Univ, the frame is the per-pixel median of the recording, which removes the walking pedestrians.
SDD is drawn in pixels on the reference frame of its scene from OpenTraj.
The soccer pitch is drawn as 105$\times$68m centered in the camera frame, since the data gives no pitch markings.

\subsection{Evaluation Protocol}
\label{sec:supp_eval}
With the error $d_{k,i}$ of hypothesis $k$ for agent $i$ (\mainpaper{sec:background}{Sec.~3.1}) and a set $\mathcal{G}$ of groups of agents, the group-level metrics match every group to its best hypothesis:
\begin{equation}
  \mathcal{G}\text{-min}_{K}\mathrm{ADE}=\frac{1}{|\mathcal{G}|}\sum_{g\in\mathcal{G}}\min_{k}\frac{1}{|g|}\sum_{i\in g}d_{k,i}.
\end{equation}
FDE is defined likewise with the error at the last frame, and min$_K$ADE is the case of single-agent groups.
The three joint metrics differ in their groups:
\textbf{G3} pairs every agent, the ball included, with its two nearest agents at the last observed frame;
\textbf{Team} takes the players of each team, so the ball counts only in G3 and Scene;
\textbf{Scene} puts all agents in one group.
The released evaluation code of MoFlow selects one hypothesis index per batch of 1,000 scenes for its joint metrics, so we evaluate the 20 hypotheses of all models with the per-scene definition above.

\subsection{Training Settings}
\label{sec:supp_settings}

\begin{table}[t]
  \centering
  \caption{Settings of H\textsuperscript{2}Flow on the team sports.}
  \label{tab:supp_hparams}
  \footnotesize
  \setlength{\tabcolsep}{4pt}
  \begin{adjustbox}{max width=\columnwidth}
  \begin{tabular}{@{}lccc@{}}
    \toprule
    & NBA & NFL & Soccer \\
    \midrule
    Agents $N$ & 11 & 23 & 23 \\
    Observed / future frames & 10 / 20 & 10 / 20 & 10 / 20 \\
    Hidden dimension $D$, heads & 128, 8 & 128, 8 & 128, 8 \\
    Encoder layers $L$, decoder blocks $B$ & 4, 4 & 4, 4 & 4, 4 \\
    Group sizes $\mathcal{S}$, segments $M$ & $\{2,3,5\}$, 2 & $\{2,3,5\}$, 2 & $\{2,3,5\}$, 2 \\
    Hyperedges per segment $E$ & 33 & 69 & 69 \\
    Temperature $\kappa$ & 0.1 & 0.1 & 0.1 \\
    Hypotheses $K$, Euler steps & 20, 10 & 20, 10 & 20, 10 \\
    Learning rate, weight decay & $10^{-3}$, 0.05 & $10^{-3}$, 0.05 & $10^{-3}$, 0.05 \\
    Batch size (scenes) & 192 & 96 & 96 \\
    Epochs $J$ & 100 & 100 & 10 \\
    Initial weight $\lambda_0$ & 0.25 & 0.25 & 0.25 \\
    \bottomrule
  \end{tabular}
  \end{adjustbox}
\end{table}

\noindent\textbf{Team Sports.}
\cref{tab:supp_hparams} lists the settings on the three sports.
On NFL and soccer, MoFlow and H\textsuperscript{2}Flow use batches of 96 scenes.
On soccer, both overfit after 2--3 epochs, so both are trained for 10 epochs and evaluated after every epoch.

\noindent\textbf{Pedestrian Benchmarks.}
On ETH-UCY and SDD, we use a learning rate of $10^{-4}$, a weight decay of 0.01, 32 or 48 agents per batch, and per-agent rotation, with pair geometry in the scene frame.
On Hotel and Zara1, a scene holds an agent and its three nearest neighbors, and the model has two pairwise interaction layers, dropout 0.1, and $\lambda_0=0.1$ and trains for 60 epochs.
On the other datasets, a scene holds one agent, and the model trains for 100--150 epochs, with the agent encoder of MoFlow on ETH and Zara2.
For the joint metrics (\cref{sec:supp_ped_joint}) and the ETH-UCY figures, separate scene models take all pedestrians of a sequence, with a learning rate of $2\times10^{-4}$, 64 agents per batch, and 100 epochs.

\subsection{Baselines}
\label{sec:supp_baselines}

\noindent\textbf{MoFlow.}
On the team sports, every result of \suppapp{this document}{this appendix} uses MoFlow retrained with chained decoder blocks and our training recipe (\mainpaper{sec:impl}{Sec.~4.2}).
Only the figures that plot the values of \mainpaper{tab:sports}{Tab.~1} (\cref{fig:supp_horizon}a and \cref{fig:supp_gains,fig:supp_cost}) and the NBA timing of \cref{tab:supp_cost} use the NBA model of that table, trained with its released code.
On NBA, the retrain is slightly more accurate than the model of \suppapp{Tab.~1}{\cref{tab:sports}}.
On ETH-UCY, MoFlow is the retrained model of \mainpaper{tab:ped}{Tab.~3}.

\noindent\textbf{Team Sports.}
We retrain LED~\cite{mao2023led}, GroupNet~\cite{xu2022groupnet}, MART~\cite{lee2024mart}, and Social-Mamba~\cite{luan2026socialmamba} with their released code on the scenes of \mainpaper{tab:sports}{Tab.~1}, with one run of 100 epochs each (10 epochs for Social-Mamba on soccer).
GroupNet forecasts 20 future frames with the agents and teams of each sport and with groups of 5 and of all agents, and MART uses its NBA configuration on all sports.
Social-Mamba forecasts every agent as the primary agent, and its joint metrics take the hypotheses with the same index.
Since the code of LED trains only its leapfrog initializer (LED-I), we first train its denoising model with its own loss, with 100 epochs for each stage.
All baselines are evaluated with our evaluation code (\cref{sec:supp_eval}), which reproduces the per-agent errors reported by the code of GroupNet and MART.

\noindent\textbf{ETH-UCY Figures.}
In the ETH-UCY figures of \cref{sec:supp_qual}, H\textsuperscript{2}Flow is the scene model of \cref{sec:supp_settings}, and MoFlow and Social-Mamba are the retrained models of \mainpaper{tab:ped}{Tab.~3}.
SingularTraj~\cite{bae2024singulartrajectory} and TUTR~\cite{shi2023tutr} are retrained with their released code and configurations (one run each), and EigenTraj~\cite{bae2023eigentrajectory} is the released pretrained model with the LB-EBM predictor, as in \mainpaper{tab:ped}{Tab.~3}.

\section{Additional Quantitative Results}
\label{sec:supp_quant}
This section extends the quantitative results (\mainpaper{sec:quant}{Sec.~4.4}), the ablation (\suppapp{Sec.~4.6}{\cref{sec:ablation}}), and the top-scored hypotheses of \suppapp{Tab.~7}{\cref{tab:sens}} (\suppapp{Sec.~4.8}{\cref{sec:sensitivity}}).

\subsection{Error over the Horizon and the Sports}
\label{sec:supp_horizon}

\begin{figure}[t]
  \centering
  \includegraphics[width=\linewidth]{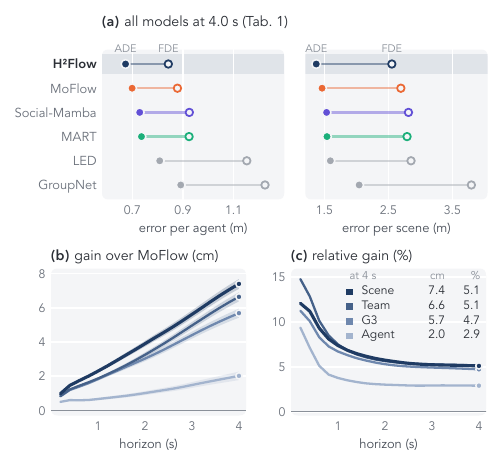}
  \caption{NBA. (a) min$_{20}$ADE (filled) and min$_{20}$FDE (open) of all models at 4.0s, per agent and per scene. (b) Error reduction of H\textsuperscript{2}Flow over MoFlow across the horizon, with 95\% intervals over the 12,500 test scenes; (c) the same, relative to MoFlow.}
  \label{fig:supp_horizon}
\end{figure}

\begin{figure}[t]
  \centering
  \includegraphics[width=\linewidth]{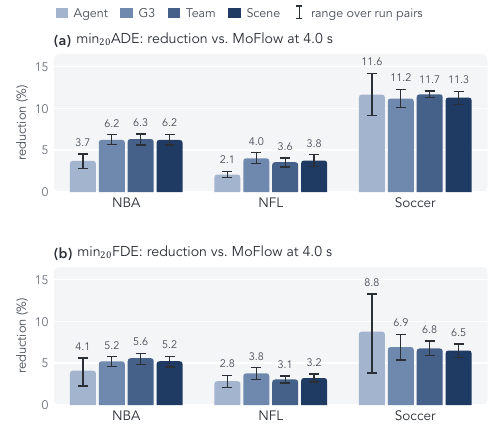}
  \caption{Error reduction of H\textsuperscript{2}Flow over MoFlow at 4.0s in (a) min$_{20}$ADE and (b) min$_{20}$FDE, per agent and for the best joint hypothesis per G3 group, team, and scene. Bars: means of \mainpaper{tab:sports}{Tab.~1}; whiskers: range over all pairs of runs.}
  \label{fig:supp_gains}
\end{figure}

\noindent\textbf{The Gains Grow with the Horizon.}
On NBA, the error reduction of H\textsuperscript{2}Flow over MoFlow grows over the horizon, to 2.0\,cm per agent and 5.7--7.4\,cm for the joint metrics at 4.0s (\cref{fig:supp_horizon}b, c).
It holds on all three sports and in every pair of runs (\cref{fig:supp_gains}).
On NFL and soccer, H\textsuperscript{2}Flow has the lowest min$_{20}$FDE from 2.0s on (\cref{tab:supp_horizon}).

\begin{table}[t]
  \centering
  \caption{Per-agent min$_{20}$ADE/FDE (m) on NFL and soccer at 1.0--4.0s; MoFlow and H\textsuperscript{2}Flow: mean of two runs. Best in bold, second underlined.}
  \label{tab:supp_horizon}
  \footnotesize
  \setlength{\tabcolsep}{4pt}
  \begin{adjustbox}{max width=\columnwidth}
  \begin{tabular}{lcccc}
    \toprule
    Method & 1.0s & 2.0s & 3.0s & 4.0s \\
    \midrule
    \multicolumn{5}{l}{\emph{NFL}} \\
    GroupNet~\cite{xu2022groupnet} & 0.39/0.61 & 0.82/1.50 & 1.36/2.47 & 1.92/3.30 \\
    LED~\cite{mao2023led} & 0.22/0.36 & 0.55/1.00 & 0.98/1.78 & 1.47/2.65 \\
    MART~\cite{lee2024mart} & 0.25/0.35 & 0.51/0.80 & 0.84/1.36 & 1.21/1.94 \\
    MoFlow~\cite{fu2025moflow} & \underline{0.21}/\textbf{0.31} & \textbf{0.44}/\underline{0.72} & \underline{0.75}/\underline{1.24} & \underline{1.10}/\underline{1.79} \\
    Social-Mamba~\cite{luan2026socialmamba} & \textbf{0.20}/\textbf{0.31} & 0.46/0.79 & 0.79/1.36 & 1.17/1.92 \\
    \cmidrule{1-5}
    \oursrow H\textsuperscript{2}Flow & 0.23/\underline{0.32} & \underline{0.45}/\textbf{0.70} & \textbf{0.74}/\textbf{1.20} & \textbf{1.08}/\textbf{1.74} \\
    \midrule
    \multicolumn{5}{l}{\emph{Soccer}} \\
    GroupNet~\cite{xu2022groupnet} & 0.21/0.33 & 0.45/0.82 & 0.76/1.41 & 1.10/2.02 \\
    LED~\cite{mao2023led} & \underline{0.20}/0.31 & 0.44/0.77 & 0.73/1.30 & 1.06/1.89 \\
    MART~\cite{lee2024mart} & \underline{0.20}/\underline{0.30} & \underline{0.43}/0.72 & 0.70/1.20 & 1.00/1.76 \\
    MoFlow~\cite{fu2025moflow} & 0.28/0.38 & 0.51/0.79 & 0.79/1.26 & 1.09/1.77 \\
    Social-Mamba~\cite{luan2026socialmamba} & \textbf{0.18}/\textbf{0.27} & \textbf{0.39}/\underline{0.68} & \textbf{0.65}/\underline{1.15} & \textbf{0.94}/\underline{1.64} \\
    \cmidrule{1-5}
    \oursrow H\textsuperscript{2}Flow & 0.24/\underline{0.30} & \underline{0.43}/\textbf{0.67} & \underline{0.68}/\textbf{1.12} & \underline{0.96}/\textbf{1.61} \\
    \bottomrule
  \end{tabular}
  \end{adjustbox}
\end{table}

\subsection{Significance of the Gains}
\label{sec:supp_signif}

\begin{table}[t]
  \centering
  \caption{Error reduction (\%) of H\textsuperscript{2}Flow relative to MoFlow at 4.0s on the full test sets, with 95\% paired bootstrap intervals over scenes, and the share of scenes where H\textsuperscript{2}Flow is lower. Wilcoxon signed-rank $p<10^{-25}$ throughout.}
  \label{tab:supp_signif}
  \footnotesize
  \setlength{\tabcolsep}{2.5pt}
  \begin{adjustbox}{max width=\columnwidth}
  \begin{tabular}{lcccc}
    \toprule
    & min$_{20}$ADE & G3 & Team & Scene \\
    \midrule
    NBA & 2.9 [2.6, 3.3] & 4.7 [4.5, 4.9] & 5.1 [4.9, 5.3] & 5.1 [4.9, 5.3] \\
    \quad scenes lower & 57\% & 67\% & 69\% & 69\% \\
    NFL & 2.4 [1.8, 3.1] & 4.6 [4.2, 5.0] & 3.9 [3.6, 4.3] & 4.2 [3.8, 4.6] \\
    \quad scenes lower & 58\% & 69\% & 66\% & 66\% \\
    Soccer & 12.4 [10.8, 14.1] & 10.7 [9.7, 11.6] & 11.4 [10.5, 12.2] & 11.2 [10.3, 12.0] \\
    \quad scenes lower & 73\% & 81\% & 85\% & 82\% \\
    \bottomrule
  \end{tabular}
  \end{adjustbox}
\end{table}

The gains of H\textsuperscript{2}Flow over MoFlow are significant on every sport and metric (\cref{tab:supp_signif}): paired over the test scenes, the 95\% bootstrap intervals of the reduction exclude zero.
H\textsuperscript{2}Flow is more accurate on most scenes, up to 85\% of them on soccer.

\subsection{Joint Accuracy on ETH-UCY}
\label{sec:supp_ped_joint}
\begin{table}[t]
  \centering
  \caption{Joint accuracy on ETH-UCY at 4.8s: JADE/JFDE (m) with all pedestrians of a test sequence in one scene (one run each). Best in bold.}
  \label{tab:supp_ped_joint}
  \footnotesize
  \setlength{\tabcolsep}{2.5pt}
  \begin{adjustbox}{max width=\columnwidth}
  \begin{tabular}{lcccccc}
    \toprule
    Method & ETH & Hotel & Univ & Zara1 & Zara2 & AVG \\
    \midrule
    MoFlow~\cite{fu2025moflow} & \textbf{.46}/\textbf{.73} & .26/.49 & \textbf{.57}/\textbf{1.18} & .32/.65 & .29/.63 & \textbf{.38}/.73 \\
    \oursrow H\textsuperscript{2}Flow & .52/.79 & \textbf{.20}/\textbf{.35} & .59/1.22 & \textbf{.30}/\textbf{.62} & \textbf{.28}/\textbf{.59} & \textbf{.38}/\textbf{.71} \\
    \bottomrule
  \end{tabular}
  \end{adjustbox}
\end{table}

JADE/JFDE match one hypothesis index to all pedestrians of a test sequence, as Scene-min$_{20}$ADE/FDE does for a play (\cref{tab:supp_ped_joint}).
H\textsuperscript{2}Flow uses the scene models of \cref{sec:supp_settings}, and MoFlow its retrained model of \mainpaper{tab:ped}{Tab.~3}.
H\textsuperscript{2}Flow has the lower JADE/JFDE on Hotel, Zara1, and Zara2 and the lower average JFDE.

\subsection{Additional Ablations}
\label{sec:supp_ablation}

\begin{table*}[t]
  \centering
  \caption{Ablation at 4.0s (m) with the variants of \mainpaper{tab:ablation}{Tab.~4}. NBA from that table, H\textsuperscript{2}Flow on soccer from \mainpaper{tab:sports}{Tab.~1} (mean of two runs), all other cells one run. Best in bold, second underlined, per sport in (a) and (b).}
  \label{tab:supp_ablation}
  \footnotesize
  \setlength{\tabcolsep}{3pt}
  \begin{adjustbox}{max width=\textwidth}
  \begin{tabular}{@{}lccccccccccccccc@{}}
    \toprule
    \multicolumn{4}{@{}l}{\emph{(a) Groups shape futures}} & \multicolumn{4}{c}{NBA} & \multicolumn{4}{c}{NFL} & \multicolumn{4}{c}{Soccer} \\
    \cmidrule(lr){5-8}\cmidrule(lr){9-12}\cmidrule(l){13-16}
    Variant & GT & GF & $\mathcal{L}_{\mathrm{grp}}$ & min$_{20}$ADE/FDE$\downarrow$ & G3$\downarrow$ & Team$\downarrow$ & Scene$\downarrow$ & min$_{20}$ADE/FDE$\downarrow$ & G3$\downarrow$ & Team$\downarrow$ & Scene$\downarrow$ & min$_{20}$ADE/FDE$\downarrow$ & G3$\downarrow$ & Team$\downarrow$ & Scene$\downarrow$ \\
    \midrule
    \oursrow H\textsuperscript{2}Flow & \checkmark & \checkmark & \checkmark & \textbf{0.672}/\textbf{0.840} & \underline{1.144} & \underline{1.226} & \underline{1.373} & \textbf{1.083}/\underline{1.761} & \textbf{1.815} & 2.279 & 2.410 & 0.964/1.611 & 1.534 & 1.588 & 1.799 \\
    w/o GT &  & \checkmark & \checkmark & 0.678/0.854 & 1.148 & 1.233 & 1.377 & \underline{1.087}/\textbf{1.737} & \underline{1.817} & 2.280 & 2.413 & 1.006/1.762 & 1.565 & 1.598 & 1.822 \\
    w/o GF & \checkmark &  & \checkmark & 0.679/0.858 & \textbf{1.139} & \textbf{1.223} & \textbf{1.365} & 1.093/1.768 & 1.819 & \textbf{2.272} & \textbf{2.403} & 0.995/1.714 & \underline{1.533} & 1.583 & \underline{1.796} \\
    w/o $\mathcal{L}_{\mathrm{grp}}$ & \checkmark & \checkmark &  & 0.692/0.871 & 1.227 & 1.307 & 1.464 & 1.090/1.764 & 1.856 & 2.330 & 2.467 & \underline{0.952}/\underline{1.577} & \textbf{1.519} & \textbf{1.545} & \textbf{1.766} \\
    w/o GT, GF &  &  & \checkmark & 0.681/0.854 & \underline{1.144} & \underline{1.226} & \underline{1.373} & 1.097/1.780 & \textbf{1.815} & \underline{2.274} & \underline{2.407} & 0.999/1.721 & 1.557 & 1.599 & 1.813 \\
    w/o GT, $\mathcal{L}_{\mathrm{grp}}$ &  & \checkmark &  & 0.690/0.869 & 1.226 & 1.309 & 1.466 & 1.098/1.780 & 1.866 & 2.338 & 2.478 & \textbf{0.949}/\textbf{1.547} & 1.534 & \underline{1.578} & 1.804 \\
    w/o GF, $\mathcal{L}_{\mathrm{grp}}$ & \checkmark &  &  & \underline{0.675}/\underline{0.848} & 1.189 & 1.282 & 1.437 & 1.098/1.773 & 1.868 & 2.343 & 2.484 & 0.961/1.603 & 1.548 & 1.602 & 1.828 \\
    \midrule
    \multicolumn{4}{@{}l}{\emph{(b) Futures define groups}} & & & & & & & & & & & & \\
    Variant & SC & FUT & TW & & & & & & & & & & & & \\
    \midrule
    \oursrow H\textsuperscript{2}Flow & \checkmark & \checkmark & \checkmark & \textbf{0.672}/\underline{0.840} & 1.144 & \underline{1.226} & 1.373 & \underline{1.083}/1.761 & \underline{1.815} & 2.279 & 2.410 & \textbf{0.964}/\textbf{1.611} & \textbf{1.534} & \underline{1.588} & \underline{1.799} \\
    w/o TW & \checkmark & \checkmark &  & \textbf{0.672}/\textbf{0.839} & 1.145 & 1.230 & 1.375 & 1.087/\underline{1.739} & 1.823 & 2.285 & 2.415 & 0.995/\underline{1.640} & 1.576 & 1.631 & 1.861 \\
    w/o FUT & \checkmark &  &  & \underline{0.673}/0.849 & \textbf{1.141} & \underline{1.226} & \underline{1.372} & \textbf{1.082}/\textbf{1.724} & 1.817 & \underline{2.275} & \underline{2.407} & 0.998/1.699 & 1.587 & 1.632 & 1.872 \\
    w/o SC &  &  &  & 0.677/0.850 & \underline{1.143} & \textbf{1.223} & \textbf{1.366} & 1.092/1.783 & \textbf{1.810} & \textbf{2.262} & \textbf{2.394} & \underline{0.988}/1.667 & \underline{1.542} & \textbf{1.578} & \textbf{1.798} \\
    \bottomrule
  \end{tabular}
  \end{adjustbox}
\end{table*}

\begin{figure*}[t]
  \centering
  \includegraphics[width=\textwidth]{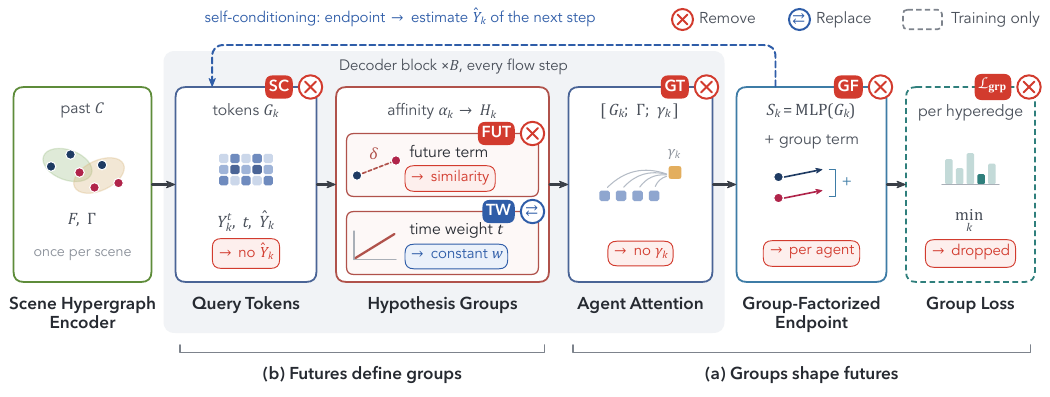}
  \caption{Where each ablated part acts in H\textsuperscript{2}Flow: rings mark the removed parts with their abbreviations in \cref{tab:supp_ablation}, and the dashed card marks the parts that act only in training. Removing SC also removes FUT and TW, and removing FUT removes TW.}
  \label{fig:supp_ablation_pipeline}
\end{figure*}

\cref{tab:supp_ablation} repeats the ablation of \mainpaper{tab:ablation}{Tab.~4} on NFL and soccer beside its NBA results, and \cref{fig:supp_ablation_pipeline} marks where each part acts.
The variants of (a) remove the parts through which the groups shape the futures: the hypothesis group tokens (GT), the group-factorized endpoint (GF), and $\mathcal{L}_{\mathrm{grp}}$.
Those of (b) remove the parts through which the futures define the groups: self-conditioning (SC), the future term (FUT), and its time weight (TW).

\noindent\textbf{The Full Model Is the Most Consistent.}
No variant is best on every sport and metric, but the full model stays closest to the best variant across all three sports, whereas every variant raises at least one of its errors by 0.05m or more.
On NFL and soccer, every variant remains more accurate than MoFlow (\mainpaper{tab:sports}{Tab.~1}).

\noindent\textbf{Each Sport Relies on Different Parts.}
As on NBA, $\mathcal{L}_{\mathrm{grp}}$ drives the joint accuracy on NFL: without it, the G3, Team, and Scene errors rise by 0.04--0.07m.
On soccer, the future term drives the joint accuracy, and the group tokens and the group-factorized endpoint lower the per-agent error.

\begin{table}[t]
  \centering
  \caption{Futures define groups without the endpoint and $\mathcal{L}_{\mathrm{grp}}$ on NBA at 4.0s (m): each ``+'' row also removes its part from the row above (SEG: the two segments of the future term). These are the variants of \suppapp{Tab.~6a of the main paper}{\cref{tab:groups}a}; the first two rows are from \suppapp{its Tab.~4}{\cref{tab:ablation}}. Best in bold, second underlined.}
  \label{tab:supp_ablation_chain}
  \footnotesize
  \setlength{\tabcolsep}{3pt}
  \begin{adjustbox}{max width=\columnwidth}
  \begin{tabular}{@{}lcccc@{}}
    \toprule
    & Per agent & \multicolumn{3}{c}{Per group (best hypothesis)} \\
    \cmidrule(lr){2-2}\cmidrule(l){3-5}
    Variant & min$_{20}$ADE/FDE$\downarrow$ & G3$\downarrow$ & Team$\downarrow$ & Scene$\downarrow$ \\
    \midrule
    \oursrow H\textsuperscript{2}Flow & \textbf{0.672}/\textbf{0.840} & \textbf{1.144} & \textbf{1.226} & \textbf{1.373} \\
    w/o GF, $\mathcal{L}_{\mathrm{grp}}$ & \underline{0.675}/\underline{0.848} & \underline{1.189} & \underline{1.282} & \underline{1.437} \\
    \hspace{0.8em}+ w/o SEG, TW & 0.682/0.858 & 1.205 & 1.293 & 1.451 \\
    \hspace{0.8em}+ w/o FUT & 0.679/0.853 & 1.204 & 1.293 & 1.450 \\
    \hspace{0.8em}+ w/o SC & 0.687/0.864 & 1.223 & 1.306 & 1.465 \\
    \bottomrule
  \end{tabular}
  \end{adjustbox}
\end{table}

\noindent\textbf{Futures Define Groups Without the Endpoint and $\mathcal{L}_{\mathrm{grp}}$.}
On NBA, \cref{tab:supp_ablation_chain} removes the parts of (b) one after another from the variant without GF and $\mathcal{L}_{\mathrm{grp}}$.
Removing the segments with the time weight, the future term, and self-conditioning raises min$_{20}$ADE from 0.675 to 0.687 and Scene-min$_{20}$ADE from 1.437 to 1.465 (\mainpaper{sec:ablation}{Sec.~4.6}).

\subsection{Number of Hypotheses}
\label{sec:supp_mink}

\begin{figure}[t]
  \centering
  \includegraphics[width=\linewidth]{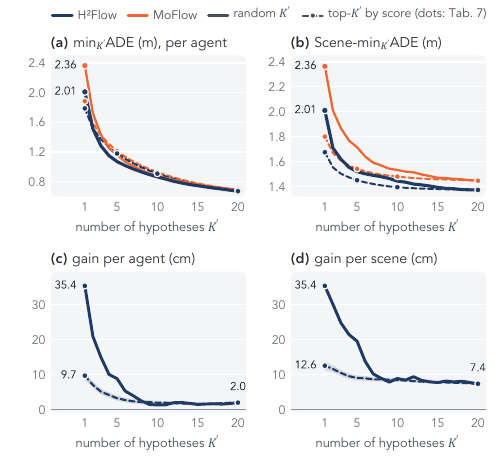}
  \caption{Best of $K'$ hypotheses on NBA. (a) min$_{K'}$ADE and (b) Scene-min$_{K'}$ADE of $K'$ random (solid; mean of 10 draws) or top-scored hypotheses (dashed; dots: the $K'$ of \mainpaper{tab:sens}{Tab.~7}). (c, d) Error reduction of H\textsuperscript{2}Flow over MoFlow, with 95\% intervals over the test scenes.}
  \label{fig:supp_mink}
\end{figure}

\cref{fig:supp_mink} extends the top-scored hypotheses of \mainpaper{tab:sens}{Tab.~7} to every $K'$.

\noindent\textbf{H\textsuperscript{2}Flow Is More Accurate for Every Number of Hypotheses.}
With random and with top-scored hypotheses, H\textsuperscript{2}Flow is more accurate than MoFlow for every $K'$, and its gain is largest for a single hypothesis.

\noindent\textbf{Scores Rank Whole Scenes.}
The top-scored scene hypotheses are more accurate than random ones for every $K'<20$.

\subsection{Diversity of the Hypotheses}
\label{sec:supp_diversity}
\begin{table}[t]
  \centering
  \caption{Diversity of the 20 hypotheses at 4.0s, mean over runs: AVar and FVar (m$^2$) and MASD (m), with min$_{20}$ADE and Scene-min$_{20}$ADE (m) for reference, the lower in bold.}
  \label{tab:supp_diversity}
  \footnotesize
  \setlength{\tabcolsep}{3pt}
  \begin{adjustbox}{max width=\columnwidth}
  \begin{tabular}{llccccc}
    \toprule
    & Method & AVar & FVar & MASD & min$_{20}$ADE & Scene \\
    \midrule
    NBA & MoFlow & 7.63 & 18.78 & 5.52 & 0.695 & 1.456 \\
    \oursrow  & H\textsuperscript{2}Flow & 7.35 & 18.16 & 5.28 & \textbf{0.673} & \textbf{1.372} \\
    \midrule
    NFL & MoFlow & 24.55 & 65.80 & 9.54 & 1.104 & 2.499 \\
    \oursrow  & H\textsuperscript{2}Flow & 23.10 & 60.94 & 8.45 & \textbf{1.081} & \textbf{2.405} \\
    \midrule
    Soccer & MoFlow & 10.31 & 26.31 & 8.13 & 1.091 & 2.028 \\
    \oursrow  & H\textsuperscript{2}Flow & 10.19 & 25.57 & 7.79 & \textbf{0.964} & \textbf{1.799} \\
    \bottomrule
  \end{tabular}
  \end{adjustbox}
\end{table}

\cref{tab:supp_diversity} measures the diversity of the 20 hypotheses with the metrics of the evaluation code of MoFlow: AVar and FVar, the variance of the hypotheses averaged over the horizon and at 4.0s, and MASD, the largest distance between two hypotheses averaged over the horizon.
The gains of H\textsuperscript{2}Flow do not come from wider sampling: on no sport do its hypotheses spread more widely than those of MoFlow.

\section{Computational Cost}
\label{sec:complexity}
Sampling evaluates the encoder once per scene and the decoder once per Euler step; self-conditioning adds no evaluation.
The group tokens grow the agent attention from $N$ to $N+(M+1)E$ tokens per hypothesis, which raises its cost per block and step from $\mathcal{O}(KN^2D)$ to $\mathcal{O}(KN(N+(M+1)E)D)$.

\subsection{Training and Sampling Time}
\label{sec:supp_timing}
\begin{table}[t]
  \centering
  \caption{Training time per epoch and sampling time per scene (20 hypotheses, default sampler; 10 Euler steps for MoFlow and H\textsuperscript{2}Flow) on one A100; LED-I trains in two stages, summed. Lowest in bold, second underlined.}
  \label{tab:supp_cost}
  \footnotesize
  \setlength{\tabcolsep}{3pt}
  \begin{adjustbox}{max width=\columnwidth}
  \begin{tabular}{lccccccc}
    \toprule
    & & \multicolumn{3}{c}{Training (min/epoch)} & \multicolumn{3}{c}{Sampling (ms/scene)} \\
    \cmidrule(lr){3-5}\cmidrule(l){6-8}
    Method & Params (M) & NBA & NFL & Soccer & NBA & NFL & Soccer \\
    \midrule
    GroupNet~\cite{xu2022groupnet} & 3.22 & 3.17 & 2.02 & 2.92 & \underline{0.31} & \underline{0.71} & \underline{0.83} \\
    LED-I~\cite{mao2023led} & 11.20 & 23.06 & 31.19 & 43.53 & 10.41 & 21.76 & 21.75 \\
    MART~\cite{lee2024mart} & \textbf{1.57} & \underline{0.46} & \textbf{0.37} & \textbf{0.47} & \textbf{0.03} & \textbf{0.11} & \textbf{0.11} \\
    MoFlow~\cite{fu2025moflow} & 3.78 & \textbf{0.24} & \underline{0.49} & \underline{0.64} & 1.61 & 2.98 & 3.02 \\
    Social-Mamba~\cite{luan2026socialmamba} & \underline{1.91} & 14.31 & 47.13 & 65.71 & 7.17 & 31.16 & 31.15 \\
    \midrule
    \oursrow H\textsuperscript{2}Flow & 8.91 & 1.57 & 2.34 & 3.26 & 6.94 & 17.22 & 17.26 \\
    \bottomrule
  \end{tabular}
  \end{adjustbox}
\end{table}

\begin{figure*}[t]
  \centering
  \includegraphics[width=\textwidth]{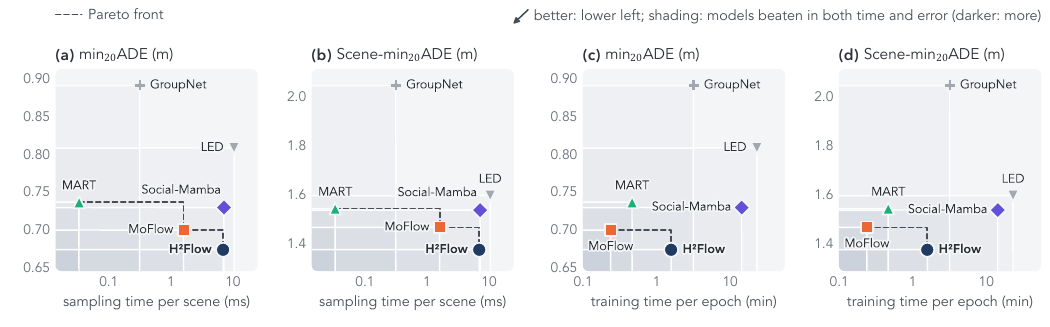}
  \caption{Accuracy against cost on NBA (\cref{tab:supp_cost}): min$_{20}$ADE and Scene-min$_{20}$ADE against (a, b) sampling time per scene and (c, d) training time per epoch. Shading: how many models a point would beat in both time and error; dashed: Pareto front.}
  \label{fig:supp_cost}
\end{figure*}

\cref{tab:supp_cost,fig:supp_cost} compare the training and sampling time of all models of \mainpaper{tab:sports}{Tab.~1} on one A100.
Training time is measured with each model's own batch size, optimizer, and loss and scaled to one epoch of each sport; sampling time is the median of five runs that draw 20 hypotheses for each of the first 1,000 test scenes (soccer: all 684).

\noindent\textbf{H\textsuperscript{2}Flow Lies on the Pareto Front.}
On NBA, for both errors, H\textsuperscript{2}Flow lies on the Pareto front of accuracy and time, in sampling and in training (\cref{fig:supp_cost}).
On all three sports, it trains and samples faster than Social-Mamba and LED (\cref{tab:supp_cost}).

\subsection{Scaling with Agents and Hypotheses}
\label{sec:supp_scaling}

\begin{figure}[t]
  \centering
  \includegraphics[width=\linewidth]{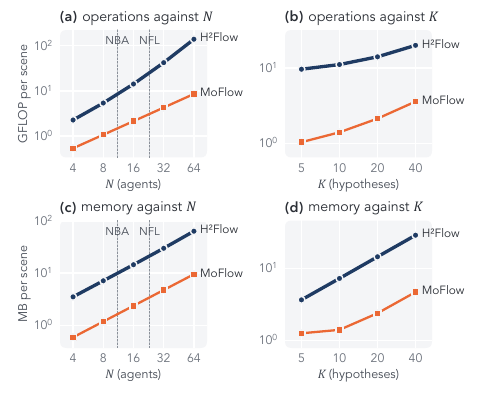}
  \caption{Sampling cost per scene (10 Euler steps, random weights, synthetic scenes) against the number of agents $N$ ($K=20$) and of hypotheses $K$ ($N=16$): (a, b) floating-point operations and (c, d) peak GPU memory. Dotted: NBA (11 agents) and NFL and soccer (23).}
  \label{fig:supp_scaling}
\end{figure}

\cref{fig:supp_scaling} measures how the sampling cost of MoFlow and H\textsuperscript{2}Flow grows with the numbers of agents and hypotheses; the operations are counted exactly, independent of the hardware.
The memory of H\textsuperscript{2}Flow grows linearly with $N$, at about 1MB per agent and scene, and the cost of both models grows linearly with $K$.

\section{Flow Steps and Hypothesis Groups}
\label{sec:supp_groups}
This section follows the sampler of \cref{alg:sample} over its Euler steps and extends \mainpaper{sec:analysis}{Sec.~4.7}.
It follows the states, endpoints (\cref{sec:supp_states}), and groups (\cref{sec:supp_flow_groups}) of the hypotheses on single plays, and gives statistics of the groups over the first 512 test plays of each sport (\cref{sec:supp_groupstats}).

\subsection{States and Endpoints}
\label{sec:supp_states}

\begin{figure*}[t]
  \centering
  \includegraphics[width=\textwidth]{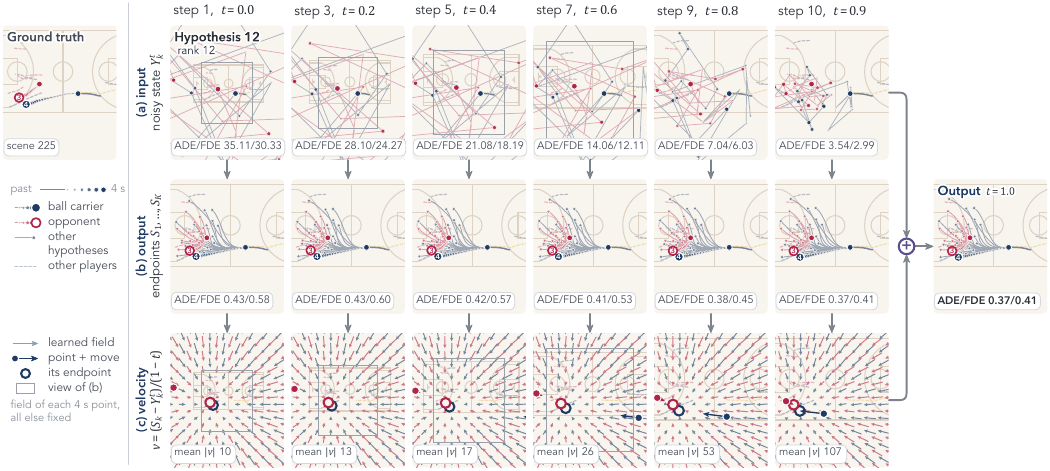}
  \caption{Flow matching on one NBA play, one column per Euler step: the ball carrier at 4.0s (navy) and its nearest opponent (crimson). (a) Noisy state $Y^t_k$ of the hypothesis best for both. (b) Endpoints $S_1,\ldots,S_K$ ($S_k$ in beads). (c) Velocity field at each player's 4.0s point, with its Euler move and endpoint. Steps 1--7 of (a, c) are zoomed out (box: view of b). Badges: ADE/FDE (m) of $S_k$; mean speed in the box.}
  \label{fig:supp_flow_process}
\end{figure*}

\cref{fig:supp_flow_process} follows two players of one play through the 10-step Euler sampler: the ball carrier at 4.0s and its nearest opponent, in the hypothesis that is best for the two.
Among the first 250 test plays with clear motion (players move 3--7m in 4.0s) that the other figures do not show, the play is the one on which the endpoint of the two players improves most from step 1 to step 10 and is most accurate at step 10.

\noindent\textbf{The States Move Straight to the Endpoints.}
The initial state is Gaussian noise shared by all $K$ hypotheses, and it spreads far beyond the court, so rows (a) and (c) zoom out before $t=0.8$.
At every step, the decoder predicts the endpoints, and the Euler step moves the state $\Delta t/(1-t)$ of the way toward them, so the error of the state falls linearly with $t$.
The last step lands on the endpoint of step 10, which is the forecast.

\noindent\textbf{The Field Points at One Endpoint.}
To draw row (c), we move the 4.0s position of a player over a grid on the court, keep the rest of the state as sampled, and evaluate the decoder at every grid point.
The field points straight at the endpoint, and its speed grows as $1/(1-t)$ while the remaining distance shrinks as $1-t$, so the points travel along straight lines at a constant speed.

\noindent\textbf{The Best Hypothesis Improves over the Steps.}
The 20 endpoints keep their spread over the steps, and the hypothesis shown, the best for the two players at every step, improves its ADE/FDE from 0.43/0.58m at step 1 to 0.37/0.41m at step 10.

\subsection{Groups over the Flow Steps}
\label{sec:supp_flow_groups}

\begin{figure*}[t]
  \centering
  \includegraphics[width=\textwidth]{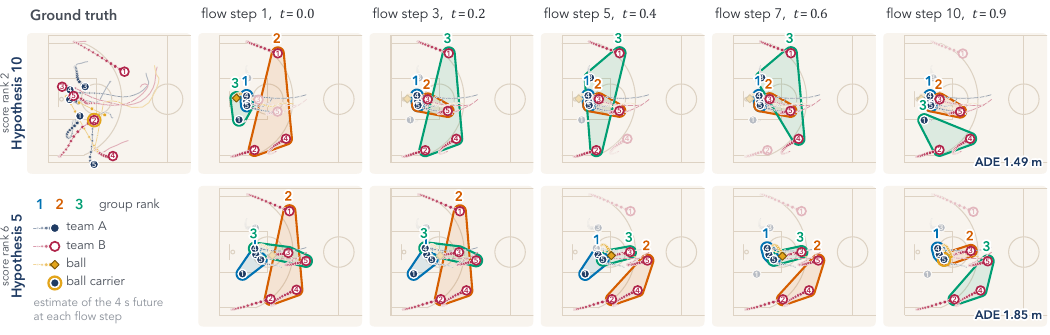}
  \caption{Estimates and groups of two hypotheses of one play at flow steps 1, 3, 5, 7, and 10.}
  \label{fig:supp_flow_groups}
\end{figure*}

\cref{fig:supp_flow_groups} follows the estimated futures and the groups of two hypotheses of one play over five flow steps.
Among the first 512 test plays, whose groups we store at every step, the play is the one with clear motion (players move more than 3.5m in 4.0s) whose 20 hypotheses form the most different 2.0--4.0s groups at the last step.
The two hypotheses are its best joint hypothesis and, among the eight top-scored others, the one whose groups and estimated future differ most from it.
Each panel shows the estimate of the 4.0s future at that step and the three groups of three agents that occur in the most hyperedges of the hypothesis in the last decoder block.

\noindent\textbf{Each Hypothesis Refines Its Groups with Its Future.}
As its estimate improves over the steps, each hypothesis adjusts its groups, and the two hypotheses keep different groups throughout.

\subsection{Statistics of the Hypothesis Groups}
\label{sec:supp_groupstats}

\begin{table}[t]
  \centering
  \caption{Hypothesis groups of H\textsuperscript{2}Flow by size, in \% (last decoder block, first 512 test scenes): groups that join both teams or hold the ball (H\textsuperscript{2}Flow / true future / random), and groups that differ between hypotheses, flow steps, or the two segments.}
  \label{tab:supp_groupstats}
  \footnotesize
  \setlength{\tabcolsep}{2.5pt}
  \begin{adjustbox}{max width=\columnwidth}
  \begin{tabular}{lcccccc}
    \toprule
    & & \multicolumn{2}{c}{Composition} & \multicolumn{3}{c}{Variation} \\
    \cmidrule(lr){3-4}\cmidrule(l){5-7}
    Sport & Size & Both teams & Ball & Hypotheses & Step & Segment \\
    \midrule
    NBA & 2 & 56 / 62 / 46 & 13 / 20 / 18 & 58 & 1.8 & 12 \\
     & 3 & 82 / 88 / 76 & 23 / 30 / 27 & 81 & 3.8 & 21 \\
     & 5 & 97 / 99 / 97 & 42 / 46 / 45 & 93 & 6.6 & 32 \\
    \midrule
    NFL & 2 & 48 / 63 / 48 & 6 / 9 / 8 & 52 & 3.6 & 23 \\
     & 3 & 73 / 83 / 75 & 9 / 12 / 13 & 72 & 6.4 & 37 \\
     & 5 & 91 / 94 / 95 & 14 / 19 / 22 & 85 & 10.0 & 49 \\
    \midrule
    Soccer & 2 & 66 / 72 / 48 & 7 / 10 / 9 & 46 & 7.6 & 15 \\
     & 3 & 87 / 86 / 76 & 10 / 15 / 13 & 64 & 12.9 & 23 \\
     & 5 & 97 / 95 / 95 & 17 / 23 / 22 & 80 & 20.2 & 33 \\
    \bottomrule
  \end{tabular}
  \end{adjustbox}
\end{table}

\cref{tab:supp_groupstats} breaks down by size the groups that H\textsuperscript{2}Flow forms, which \mainpaper{sec:analysis}{Sec.~4.7} summarizes over all sizes.

\noindent\textbf{The Groups Mix the Teams.}
On NBA and soccer, pairs and groups of three join both teams more often than random groups of the same size and nearly as often as the groups of the true future; groups of five almost always join both teams.

\noindent\textbf{Larger Groups Vary More.}
Groups of five differ more than pairs between the hypotheses of a play, between flow steps, and between the two segments of the horizon.

\section{Additional Qualitative Results}
\label{sec:supp_qual}
This section extends \mainpaper{sec:qual}{Sec.~4.5}.
\ifappendixmode  %
It first gives the rules that select the plays (\cref{sec:supp_selection}).
It then shows the best joint hypothesis of every model on NFL, soccer, and ETH-UCY with failure cases against MoFlow (\cref{sec:supp_qual_best}), the hypothesis groups (\cref{sec:supp_qual_groups}), and all 20 futures of two agents (\cref{sec:supp_qual_futures}).
\else
It first gives the rules that select the plays (\cref{sec:supp_selection}) and compares H\textsuperscript{2}Flow with MoFlow, including failure cases (\cref{sec:supp_qual_moflow}).
It then shows, on the three sports and on ETH-UCY, the best joint hypothesis of every model (\cref{sec:supp_qual_best}), the hypothesis groups (\cref{sec:supp_qual_groups}), and all 20 futures of two agents (\cref{sec:supp_qual_futures}).
\fi
ETH-UCY is drawn on the camera frames (\cref{sec:supp_data}).

\subsection{Selection of the Plays}
\label{sec:supp_selection}
Fixed rules select and rank the candidate plays and sequences, and no NBA play appears in two figures.
NBA and NFL plays come from the first 1,000 test scenes and soccer plays from all 684, except for the hypothesis groups, which use the first 512 scenes of each sport, for which the groups are stored.
Unless stated otherwise, the players move 3--7m in 4.0s on average (3--10m on NFL and soccer).
Each kind of figure has its own candidates and ranking:
\begin{itemize}
  \setlength{\itemsep}{1pt}
  \ifappendixmode\else
  \item \textbf{All 20 hypotheses} (\cref{fig:supp_samples}): plays in which the players move 1.5--3.5m, ranked by how much H\textsuperscript{2}Flow lowers Scene-min$_{20}$ADE relative to MoFlow.
  \item \textbf{Transition plays} (\cref{fig:supp_transitions}): plays in which the players move more than 4m (11--15m on the plays shown), the Scene-min$_{20}$ADE of MoFlow exceeds 1.3m, and that of H\textsuperscript{2}Flow is below 70\% of it, ranked by the ratio of the two.
  \item \textbf{Failure cases} (\cref{fig:supp_failures}): plays on which the Scene-min$_{20}$ADE of H\textsuperscript{2}Flow exceeds that of MoFlow, ranked by the ratio of the two.
  \fi
  \item \textbf{Best joint hypothesis of all models} (\mainpaper{fig:qual_pred_gt}{Fig.~3} and \suppapp{\cref{fig:supp_pred_gt,fig:supp_pred_gt_nfl,fig:supp_pred_gt_soccer}}{\cref{fig:supp_pred_gt_nfl,fig:supp_pred_gt_soccer}}): plays on which H\textsuperscript{2}Flow has the lowest Scene-min$_{20}$ADE of all models, ranked by its margin over the best other model.
  On ETH-UCY (\cref{fig:supp_pred_gt_ethucy}): test sequences of 3--8 pedestrians who each walk at least 1m in 4.8s, ranked by the ratio of the best joint ADE of H\textsuperscript{2}Flow to that of the best other model.
  \ifappendixmode
  \item \textbf{Failure cases} (\cref{fig:supp_failures}): plays on which the Scene-min$_{20}$ADE of H\textsuperscript{2}Flow exceeds that of MoFlow, ranked by the ratio of the two.
  \fi
  \item \textbf{Hypothesis groups} (\mainpaper{fig:qual_hyp_groups}{Fig.~4} and \suppapp{\cref{fig:supp_hyp_groups,fig:supp_hyp_groups_sports}}{\cref{fig:supp_hyp_groups_sports}}): plays whose best joint ADE is below the median and on which at least two thirds of the groups shown are distinct, ranked by the mean ADE of the hypotheses shown.
  The hypotheses are chosen without the ground truth: the top-scored one first, then each time the one among the ten top-scored whose groups differ most from those already chosen.
  On ETH-UCY (\cref{fig:supp_hyp_groups_ethucy}): test sequences of 6--10 pedestrians who each walk at least 1m in 4.8s, with the same distinctness condition and ranking and no bound on the joint error, which grows with the number of pedestrians.
  \item \textbf{All 20 futures of two agents} (\cref{fig:qual_k_futures,fig:supp_k_futures,fig:supp_k_futures_nfl,fig:supp_k_futures_soccer}): plays with a pass of more than 6m between teammates on which H\textsuperscript{2}Flow has the lowest error for the receiver and its nearest opponent, ranked by its margin over the better baseline.
  On ETH-UCY (\cref{fig:supp_k_futures_ethucy}): pairs of pedestrians who stay within 2m of each other, in test sequences of 3--8 pedestrians who each walk at least 1m in 4.8s, on which H\textsuperscript{2}Flow has the lowest error for the two, ranked by its ratio to that of the better baseline.
\end{itemize}
The plays of \cref{fig:supp_flow_process,fig:supp_flow_groups} follow the rules of \cref{sec:supp_states,sec:supp_flow_groups}.
\newcommand{\figfailures}{%
\begin{figure}[t]
  \centering
  \includegraphics[width=\appscale{0.88}\linewidth]{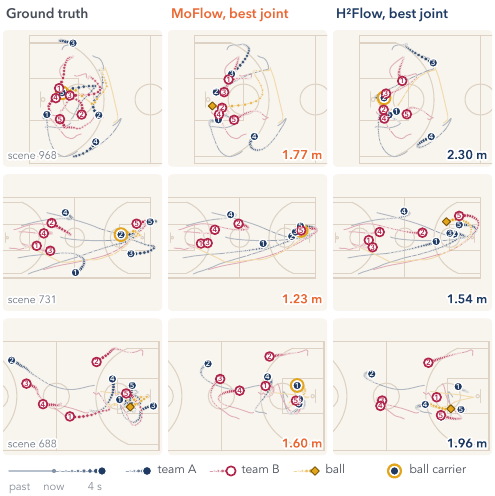}  %
  \caption{Failure cases: best joint hypothesis of MoFlow and H\textsuperscript{2}Flow on the three NBA plays on which H\textsuperscript{2}Flow errs most relative to MoFlow.}
  \label{fig:supp_failures}
\end{figure}
}
\newcommand{\textfailures}{%
\noindent\textbf{Failure Cases.}
\cref{fig:supp_failures} shows the three plays on which H\textsuperscript{2}Flow errs most relative to MoFlow.
The gap comes mostly from the ball on the first play and from three players on the second.
On the third, H\textsuperscript{2}Flow is more accurate per agent, but its best joint hypothesis errs more than that of MoFlow on two players of one team.}
\ifappendixmode\else
\subsection{Comparison with MoFlow}
\label{sec:supp_qual_moflow}

\begin{figure}[t]
  \centering
  \includegraphics[width=\linewidth]{figs/supp/fig_samples.pdf}
  \caption{All 20 hypotheses of MoFlow and H\textsuperscript{2}Flow on two NBA plays, best joint hypothesis on top.}
  \label{fig:supp_samples}
\end{figure}

\begin{figure}[t]
  \centering
  \includegraphics[width=\linewidth]{figs/supp/fig_transitions.pdf}
  \caption{Best joint hypothesis of MoFlow and H\textsuperscript{2}Flow on three NBA transition plays.}
  \label{fig:supp_transitions}
\end{figure}

\figfailures

\cref{fig:supp_samples} shows all 20 hypotheses of MoFlow and H\textsuperscript{2}Flow on two half-court plays, with the best joint hypothesis on top.
\cref{fig:supp_transitions} compares the best joint hypotheses of the two models on three transition plays, and \cref{fig:supp_failures} on three failure cases.

\textfailures
\fi

\ifappendixmode\else  %
\begin{figure*}[tp]
  \centering
  \includegraphics[width=\textwidth]{figs/supp/fig_pred_gt_more.pdf}
  \caption{Best joint hypothesis of each model on three further NBA plays.}
  \label{fig:supp_pred_gt}
\end{figure*}
\fi

\begin{figure*}[tp]
  \centering
  \includegraphics[width=\textwidth]{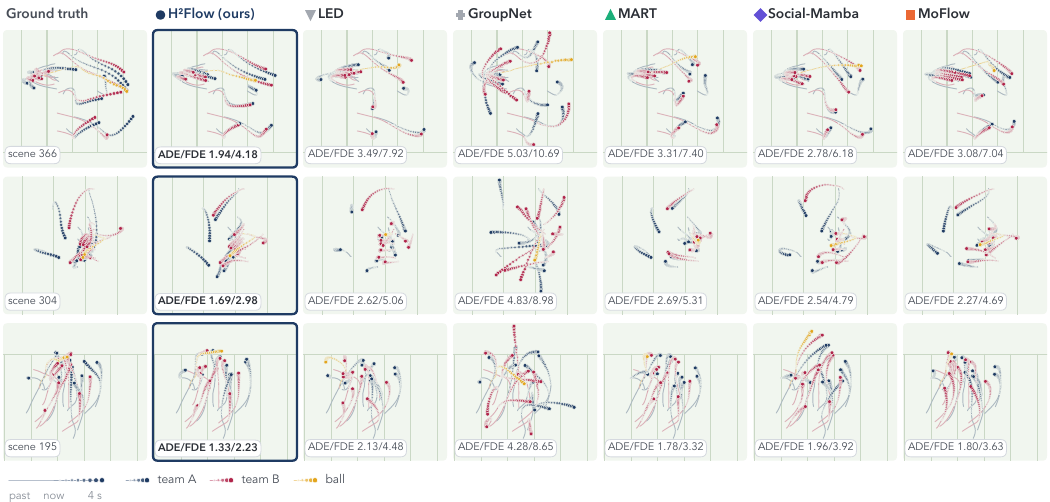}
  \caption{Best joint hypothesis of each model on three NFL plays.}
  \label{fig:supp_pred_gt_nfl}
\end{figure*}

\begin{figure*}[tp]
  \centering
  \includegraphics[width=\textwidth]{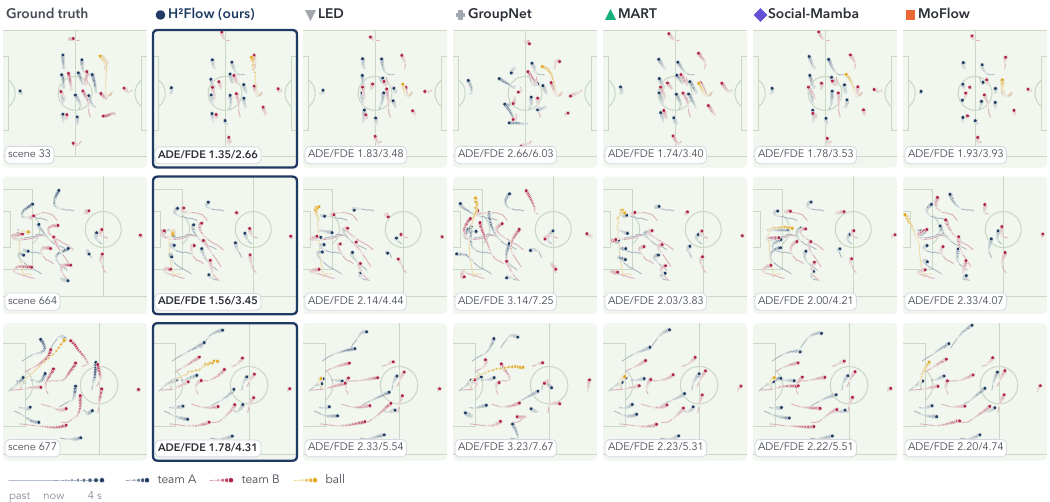}
  \caption{Best joint hypothesis of each model on three soccer plays.}
  \label{fig:supp_pred_gt_soccer}
\end{figure*}

\begin{figure*}[tp]
  \centering
  \includegraphics[width=\textwidth]{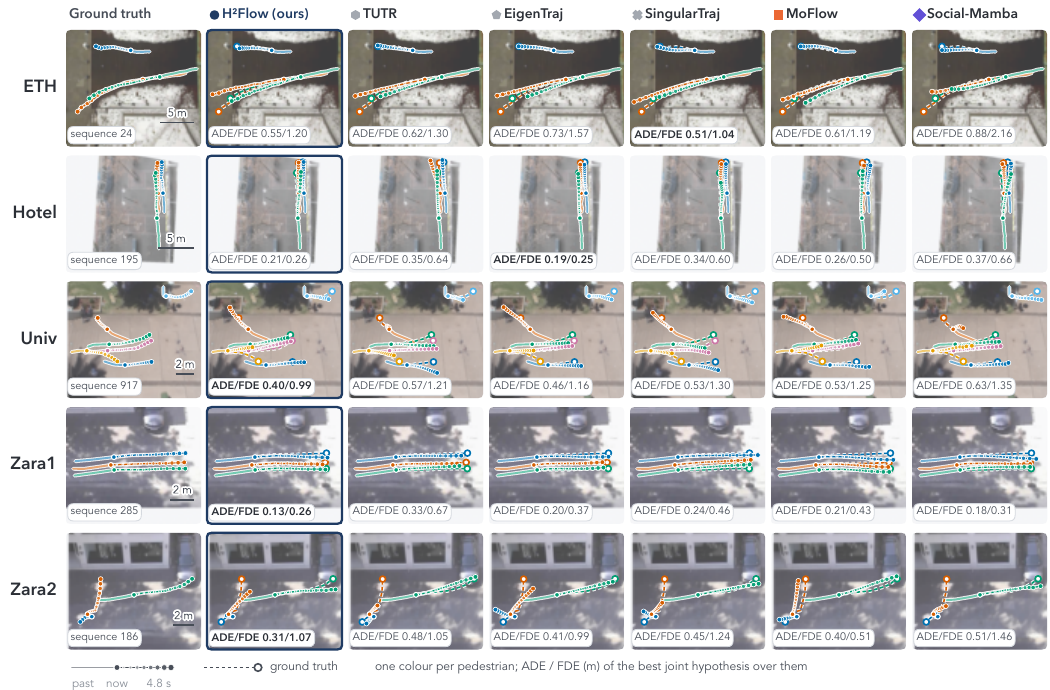}
  \caption{Best joint hypothesis of each model on one test sequence per ETH-UCY subset, on the camera frame. One color per pedestrian; dashed line ending in a ring: true future.}
  \label{fig:supp_pred_gt_ethucy}
\end{figure*}

\ifappendixmode\figfailures\fi

\ifappendixmode\else  %
\begin{figure*}[tp]
  \centering
  \includegraphics[width=\textwidth]{figs/supp/fig_hyp_groups_382.pdf}\\[4pt]
  \includegraphics[width=\textwidth]{figs/supp/fig_hyp_groups_291.pdf}
  \caption{Hypothesis groups of H\textsuperscript{2}Flow on two further NBA plays.}
  \label{fig:supp_hyp_groups}
\end{figure*}
\fi

\newcommand{\figscales}{%
\begin{figure*}[tp]
  \centering
  \includegraphics[width=\textwidth]{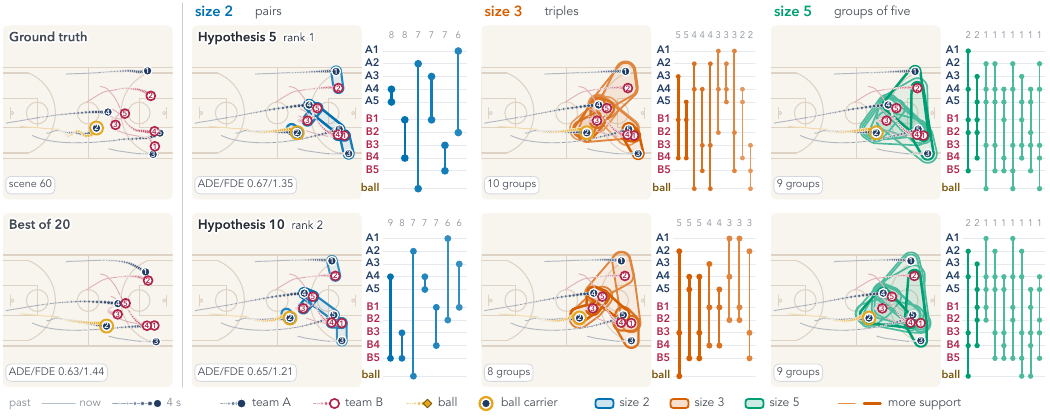}
  \caption{Every group of each size that the two top-scored hypotheses of \mainpaper{fig:qual_hyp_groups}{Fig.~4} form (2.0--4.0s, last decoder block and flow step), as hulls around the members' predicted 4.0s positions and as incidence matrices (support above each column; stronger: more support). Left: ground truth (top) and best of 20 (bottom).}
  \label{fig:supp_hyp_groups_scales}
\end{figure*}
}
\ifappendixmode\else\figscales\fi

\begin{figure*}[tp]
  \centering
  \ifappendixmode  %
  \includegraphics[width=0.85\textwidth]{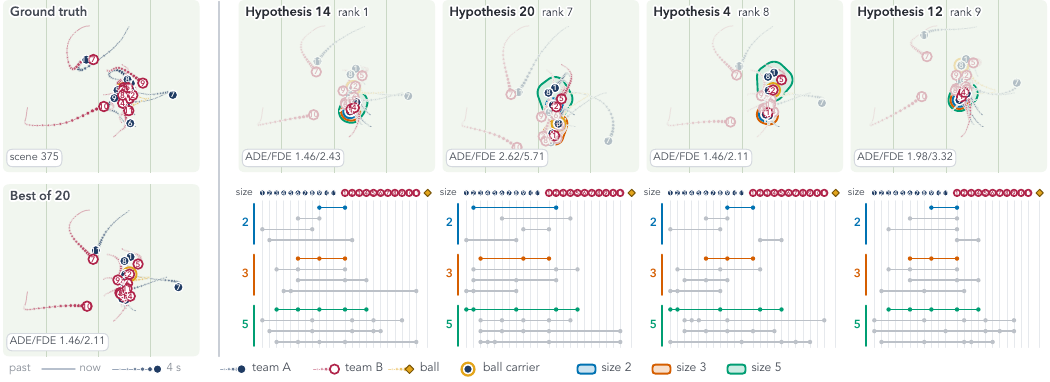}\\[4pt]
  \includegraphics[width=0.85\textwidth]{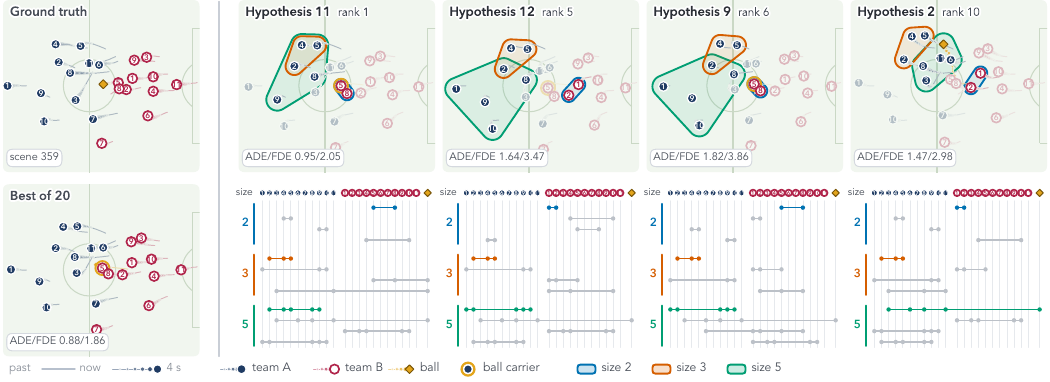}
  \else
  \includegraphics[width=\textwidth]{figs/supp/fig_hyp_groups_nfl.pdf}\\[4pt]
  \includegraphics[width=\textwidth]{figs/supp/fig_hyp_groups_soccer.pdf}
  \fi
  \caption{Hypothesis groups of H\textsuperscript{2}Flow on one NFL play (top) and one soccer play (bottom), with \suppapp{three}{four} hypotheses each.\suppapp{}{ The plays are selected with three hypotheses (\cref{sec:supp_selection}); the fourth is the next one the same rule picks, so some groups repeat.}}
  \label{fig:supp_hyp_groups_sports}
\end{figure*}

\begin{figure*}[tp]
  \centering
  \ifappendixmode  %
  \includegraphics[width=0.83\textwidth]{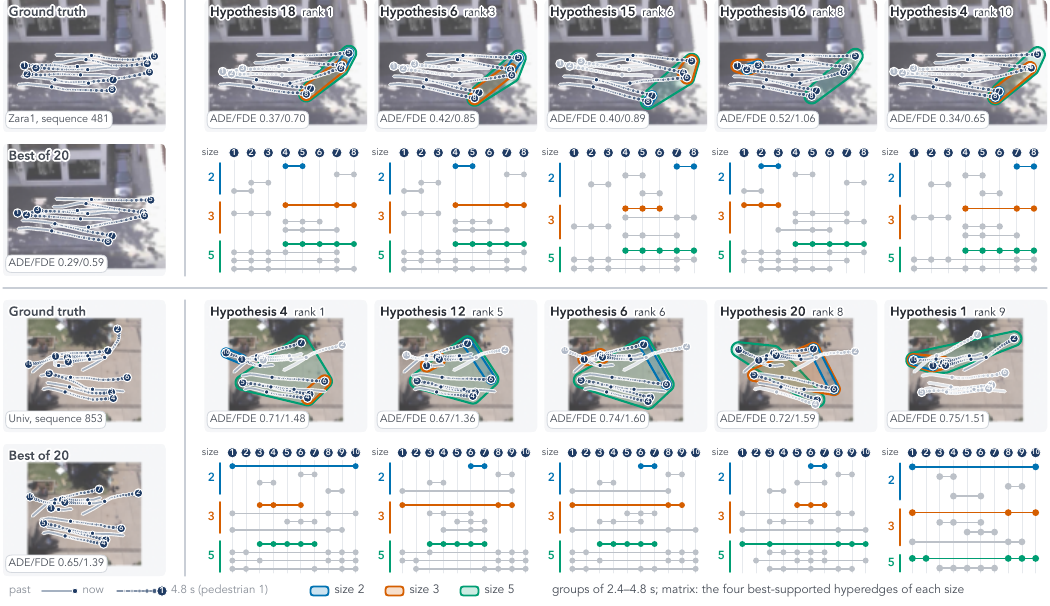}
  \else
  \includegraphics[width=\textwidth]{figs/supp/fig_hyp_groups_ethucy.pdf}
  \fi
  \caption{Hypothesis groups of H\textsuperscript{2}Flow on one Zara1 (top) and one Univ (bottom) test sequence, with \suppapp{three}{five} hypotheses each, on the camera frame.\suppapp{}{ The sequences are selected with three hypotheses; the other two are the next ones the same rule picks, so some groups repeat.} Pedestrians are numbered as the columns of the incidence matrices; hulls: the best-supported group of each size at 2.4--4.8s.}
  \label{fig:supp_hyp_groups_ethucy}
\end{figure*}

\ifappendixmode\figscales\fi

\begin{figure*}[tp]
  \centering
  \begin{minipage}[t]{\columnwidth}
    \vspace{0pt}
    \centering
    \includegraphics[width=\linewidth]{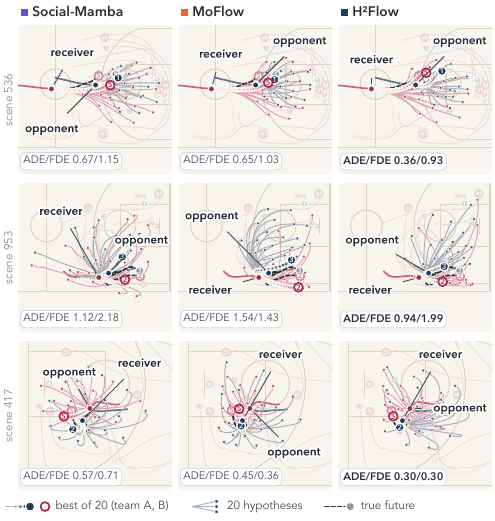}
    \caption{All 20 futures of a pass receiver and its nearest opponent on three NBA plays.}
    \label{fig:qual_k_futures}
  \end{minipage}\hfill
  \begin{minipage}[t]{\columnwidth}
    \vspace{0pt}
    \centering
    \includegraphics[width=\linewidth]{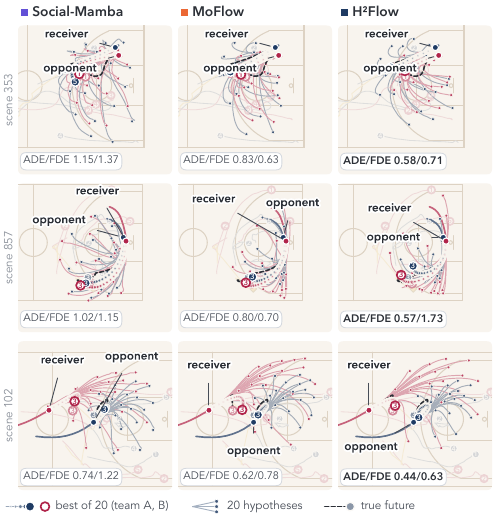}
    \caption{All 20 futures of a pass receiver and its nearest opponent on three further NBA plays.}
    \label{fig:supp_k_futures}
  \end{minipage}
\end{figure*}

\begin{figure*}[tp]
  \centering
  \begin{minipage}[t]{\columnwidth}
    \vspace{0pt}
    \centering
    \includegraphics[width=\linewidth]{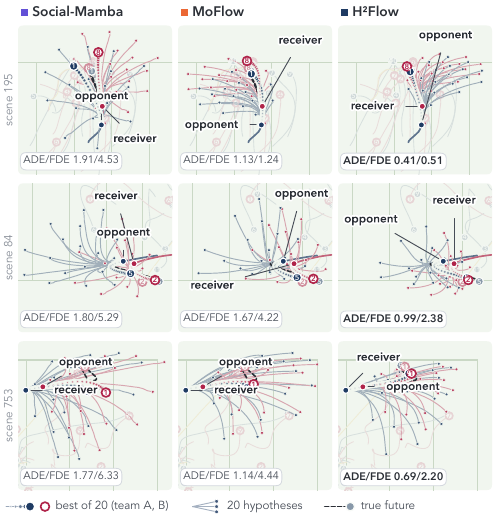}
    \caption{All 20 futures of a pass receiver and its nearest opponent on three NFL plays.}
    \label{fig:supp_k_futures_nfl}
  \end{minipage}\hfill
  \begin{minipage}[t]{\columnwidth}
    \vspace{0pt}
    \centering
    \includegraphics[width=\linewidth]{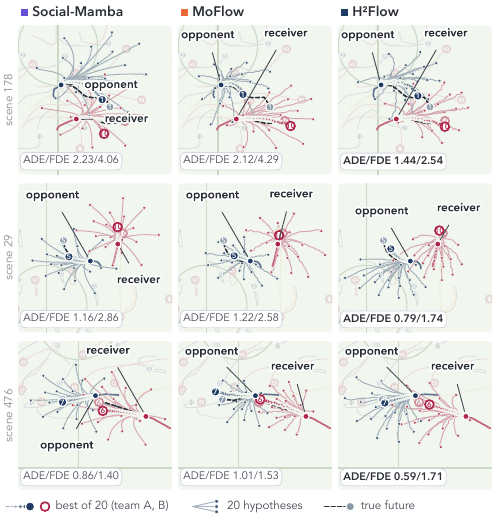}
    \caption{All 20 futures of a pass receiver and its nearest opponent on three soccer plays.}
    \label{fig:supp_k_futures_soccer}
  \end{minipage}
\end{figure*}

\begin{figure*}[tp]
  \centering
  \includegraphics[width=\textwidth]{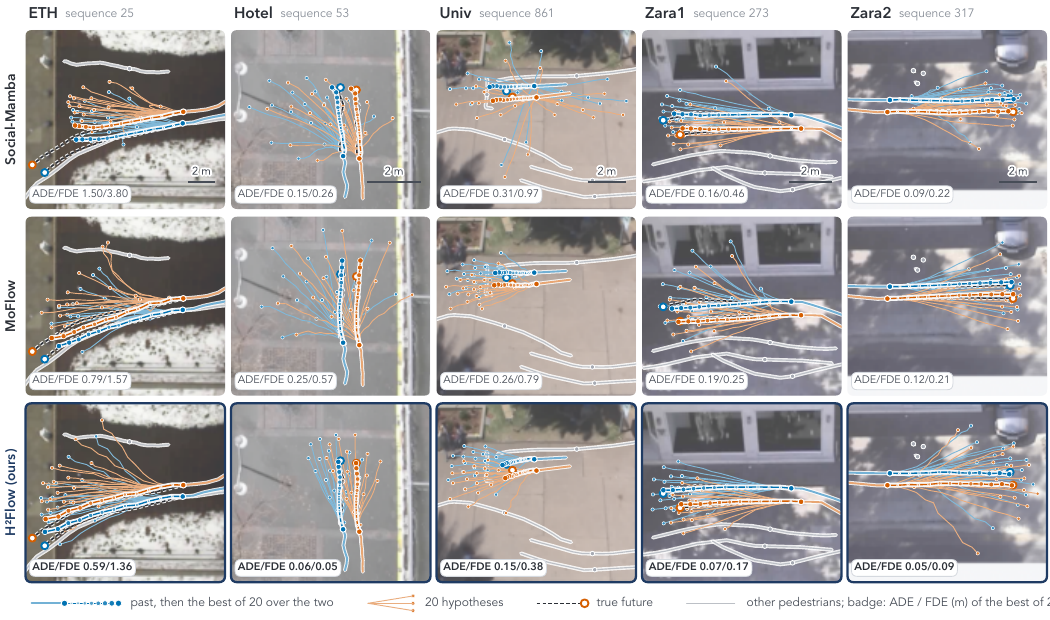}
  \caption{All 20 futures of two pedestrians who walk together on one test sequence of each ETH-UCY subset, on the camera frame: Social-Mamba, MoFlow, and H\textsuperscript{2}Flow (rows). Hairlines: the 20 futures of each pedestrian; beads: the best of the 20 for the two together, with its ADE/FDE (m) in the badge; dashed line ending in a ring: true future; gray lines: the other pedestrians.}
  \label{fig:supp_k_futures_ethucy}
\end{figure*}

\subsection{Best Joint Hypothesis of All Models}
\label{sec:supp_qual_best}
\suppapp{\cref{fig:supp_pred_gt} shows the next three NBA plays of the rule of \mainpaper{fig:qual_pred_gt}{Fig.~3}, and \cref{fig:supp_pred_gt_nfl,fig:supp_pred_gt_soccer} the three top-ranked NFL and soccer plays.}{\cref{fig:supp_pred_gt_nfl,fig:supp_pred_gt_soccer} show the three top-ranked NFL and soccer plays of the rule of \cref{fig:qual_pred_gt}.}
\cref{fig:supp_pred_gt_ethucy} shows one test sequence of each ETH-UCY subset with the six models of \cref{sec:supp_baselines}; as for JADE (\cref{tab:supp_ped_joint}), the best joint hypothesis uses one hypothesis index for all pedestrians of a sequence.
\ifappendixmode

\textfailures
\fi

\subsection{Hypothesis Groups}
\label{sec:supp_qual_groups}
\suppapp{\cref{fig:supp_hyp_groups} shows the next two NBA plays of the rule of \mainpaper{fig:qual_hyp_groups}{Fig.~4}, and \cref{fig:supp_hyp_groups_sports} one NFL and one soccer play}{\cref{fig:supp_hyp_groups_sports} shows one NFL and one soccer play of the rule of \cref{fig:qual_hyp_groups}}; with 23 agents, these show \suppapp{three}{four} hypotheses each, so that the incidence matrices stay legible.
\newcommand{\textscales}{%
\cref{fig:supp_hyp_groups_scales} shows every group of each size that the two top-scored hypotheses of \mainpaper{fig:qual_hyp_groups}{Fig.~4} form, 6--10 distinct groups per size and hypothesis, with their incidence matrices.
The group shown for each size in \suppapp{Fig.~4}{\cref{fig:qual_hyp_groups}} is the best-supported one, the first column of each matrix.}
\ifappendixmode\else\textscales\fi

\noindent\textbf{ETH-UCY.}
\cref{fig:supp_hyp_groups_ethucy} shows the groups of the ETH-UCY scene models (\cref{sec:supp_settings}) on the top-ranked Zara1 and Univ sequences, with \suppapp{three}{five} hypotheses each.
In the Zara1 sequence, pedestrians 1--3 walk left and pedestrians 4--8 walk right; all \suppapp{three}{five} hypotheses group the five who walk right, and hypothesis 16 also groups the three who walk left.
The groups change with the hypothesis: on the Univ sequences of at least four pedestrians, the 20 hypotheses form a median of 20 distinct sets of 2.4--4.8s groups.
\ifappendixmode

\noindent\textbf{All Groups of \cref{fig:qual_hyp_groups}.}
\textscales
\fi

\subsection{All Futures of Two Agents}
\label{sec:supp_qual_futures}
\cref{fig:qual_k_futures,fig:supp_k_futures} show all 20 futures of a pass receiver and its nearest opponent on the six top-ranked NBA plays of the rule, and \cref{fig:supp_k_futures_nfl,fig:supp_k_futures_soccer} on three NFL and three soccer plays.
Each panel shows the true future (dashed) and the best of the 20 futures for the two players (beads), with its ADE/FDE in the badge.

\cref{fig:supp_k_futures_ethucy} shows all 20 futures of two pedestrians who walk together on one test sequence of each ETH-UCY subset, for Social-Mamba, MoFlow, and H\textsuperscript{2}Flow.

\fi

\end{document}